\documentclass[lettersize,journal]{IEEEtran}
\IEEEoverridecommandlockouts

\usepackage{cite}
\usepackage{amsmath,amssymb,amsfonts}
\usepackage{graphicx}
\usepackage{textcomp}
\usepackage{xcolor}
\usepackage{lipsum} 
\usepackage{adjustbox}
\usepackage{booktabs}
\usepackage{wrapfig}
\usepackage{tablefootnote} % for table footnotes
\usepackage[
  bookmarks=false,
  pdfpagelabels=false,
  hyperfootnotes=false,
  hyperindex=false,
  pageanchor=false,
  citecolor=red,
  colorlinks,
]{hyperref}

\usepackage{algorithm,algpseudocode}
\usepackage{tikz}
\usepackage{xcolor}
\newcommand*\circled[1]{\tikz[baseline=(char.base)]{
            \node[shape=circle,fill,inner sep=0.8pt] (char) {\textcolor{white}{#1}};}}

\newcommand{\etal}{\textit{et al}.}
\newcommand{\ie}{\textit{i}.\textit{e}., }
\newcommand{\eg}{\textit{e}.\textit{g}., } 

\begin{document}

\title{Workload-Balanced Pruning for \\ Sparse Spiking Neural Networks
% {\footnotesize \textsuperscript{}}
}

% \author{Ruokai Yin,~Youngeun Kim,~\IEEEmembership{Student Member,~IEEE,~}\\Yuhang Li,~Abhishek Moitra,~\IEEEmembership{Student Member,~IEEE,~} \\
% Nitin Satpute,~Anna Hambitzer,~\IEEEmembership{Member,~IEEE,~}and Priyadarshini Panda,~\IEEEmembership{Member, IEEE} }

\author{Ruokai Yin,~Youngeun Kim,~Yuhang Li,~Abhishek Moitra,~\IEEEmembership{Student Member,~IEEE,~} \\
{Nitin Satpute,~Anna Hambitzer,~}and Priyadarshini Panda,~\IEEEmembership{Member, IEEE} }

\maketitle

\begin{abstract}
% \lipsum[1]
Pruning for Spiking Neural Networks (SNNs) has emerged as a fundamental methodology for deploying deep SNNs on resource-constrained edge devices. Though the existing pruning methods can provide extremely high weight sparsity for deep SNNs, the high weight sparsity brings a workload imbalance problem.
Specifically, the workload imbalance happens when a different number of non-zero weights are assigned to hardware units running in parallel. This results in low hardware utilization and thus imposes longer latency and higher energy costs.
In preliminary experiments, we show that sparse SNNs ($\sim$98\% weight sparsity) can suffer as low as $\sim$59\% utilization. To alleviate the workload imbalance problem, we propose u-Ticket, where we monitor and adjust the weight connections of the SNN during Lottery Ticket Hypothesis (LTH) based pruning, thus guaranteeing the final ticket gets optimal utilization when deployed onto the hardware. Experiments indicate that our u-Ticket can guarantee up to 100\% hardware utilization, thus reducing up to 76.9\% latency and 63.8\% energy cost compared to the non-utilization-aware LTH method.\footnote{Code is available at \color{magenta}{\href{https://github.com/Intelligent-Computing-Lab-Yale/u-Ticket-Pruning}{https://github.com/Intelligent-Computing-Lab-Yale/u-Ticket-Pruning}}}
\end{abstract}

\begin{IEEEkeywords}
Spiking Neural Networks, Pruning, Neuromorphic Computing, Sparse Neural Networks
\end{IEEEkeywords}

\section{Introduction}

Spiking Neural Networks (SNNs) have gained tremendous attention towards ultra-low-power machine learning \cite{roy2019towards}. SNNs leverage spatio-temporal information of unary spike data to achieve energy-efficient processing in resource-constrained edge devices \cite{davies2018loihi,akopyan2015truenorth}. However, in the case of large-scale tasks such as image classification, the model size of SNNs significantly increases. 
Unfortunately, edge devices typically have limited on-chip memory, rendering large-scale SNN deployment unpractical. To this end, recent works have proposed various unstructured SNN pruning techniques to achieve high weight sparsity in SNNs \cite{rathi2018stdp,neftci2016stochastic,deng2021comprehensive,guo2020unsupervised,chen2021pruning,shi2019soft,han2022adaptive,kim2022exploring}. 

Although unstructured pruning manages to compress the SNN models into the available memory resources, sparse SNNs encounter a \textbf{workload-imbalance problem} \cite{loadimbalance}. The workload-imbalance problem comes from the conventional weight stationary dataflow \cite{eyeriss} adopted in sparse accelerators \cite{ gospa,sparten,scnn}. In weight stationary dataflow, filters are divided into several groups and kept stationary inside processing elements (PEs) for filter reuse. However, different filter groups inevitably have different densities of non-zero weights due to the random weight connections from the unstructured pruning. As a result, different PEs end up with unbalanced workloads. Since all PEs run in parallel, PEs with fewer workloads must wait for the PE with the largest workload. This results in low utilization and imposes idle cycles, increasing latency and leakage energy.

\begin{figure}[t]
\centerline{\includegraphics[width=0.95\linewidth]{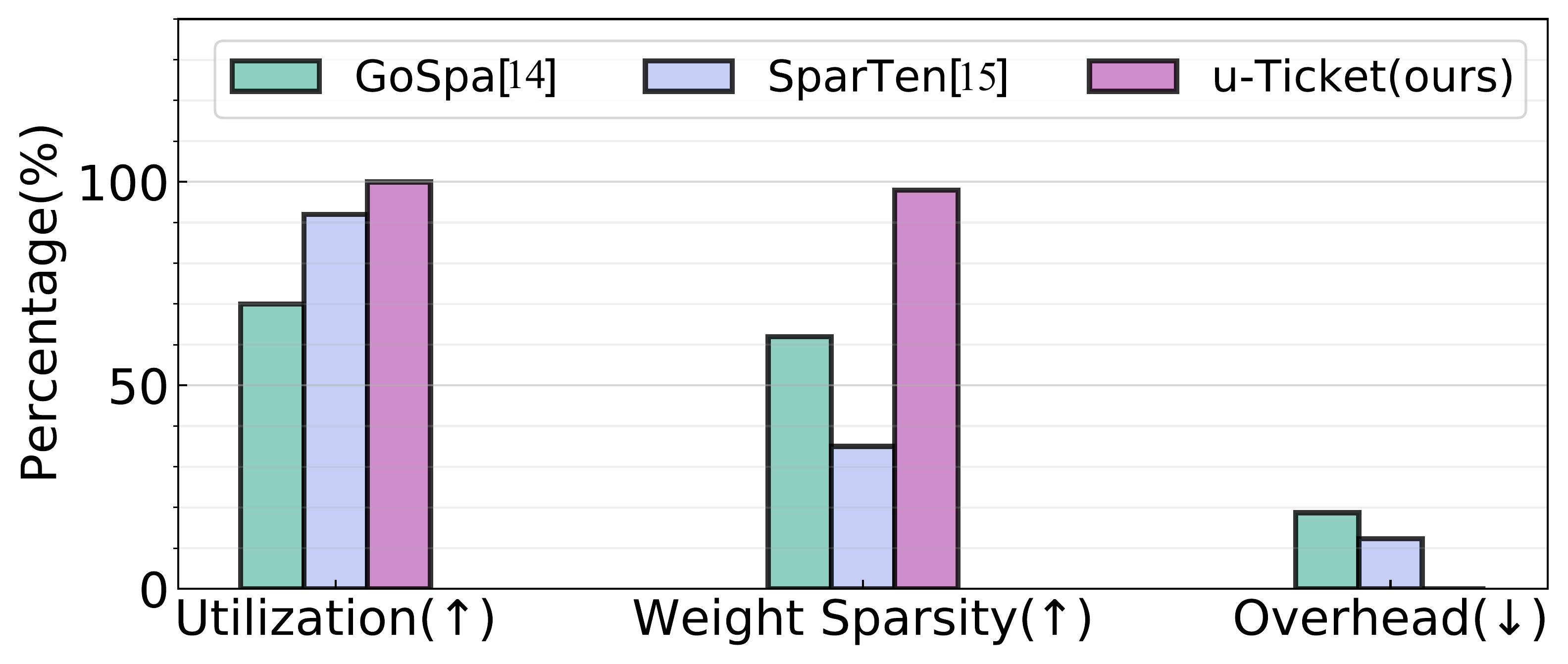}}

\caption{Comparison between u-Ticket and state-of-the-art workload balance methods. Overall, u-Ticket recovers the PE utilization up to $100\%$ for extremely sparse networks with 98\% weight sparsity (here, we consider %utilization and weight sparsity 
VGG-16). Please note that u-Ticket does not introduce any hardware area overhead and, thus, is the best fit for SNNs
(↑: the higher is the better, ↓: the lower is the better).
}

\label{intro_comp}
\end{figure}

To address the workload-imbalance problem, various methods have been proposed in the prior sparse accelerator designs.
However, they cannot be efficiently applied to SNNs for the following reasons:

\textbf{(1) {Requiring extra hardware:}} 
The prior methods require extra hardware (\eg deep FIFOs or permuting units) \cite{sparten,gospa,eie,column,ese} to balance the workloads. For instance, applying the hardware-based (FIFOs \cite{gospa} and permuting networks \cite{sparten}) workload balancing methods to SNNs require approximately 18\% and 13\% of extra chip area (see Fig. \ref{intro_comp}). Consequentially, the improvements in PE utilization are at the cost of additional hardware resources, which should be avoided for SNNs whose running environments are typically resource-constrained edge devices.

\textbf{(2) {Limited to low sparsity:}}
As shown in Fig. \ref{intro_comp}, the solutions from prior sparse accelerators \cite{gospa,sparten} only work on low sparsity (roughly 60\% and 35\% on VGG-16), which is not sufficient for SNNs' extremely low-power edge deployment. Moreover, the workload-imbalance problem naturally becomes more difficult to solve at high sparsity regimes. Hence, the exploration of workload balancing for extremely sparse networks ($>95\%$ weight sparsity) is missing in prior works. Considering the abovementioned problems, we need an SNN-friendly solution to address the workload imbalance.

\begin{figure}[t]
\begin{center}
\def\arraystretch{0.5}
\begin{tabular}{@{\hskip 0.0\linewidth}c@{\hskip 0.0\linewidth}c@{\hskip 0.00\linewidth}c@{\hskip 0.00\linewidth}c@{}c}
\hspace{-10mm}
\includegraphics[width=\linewidth]{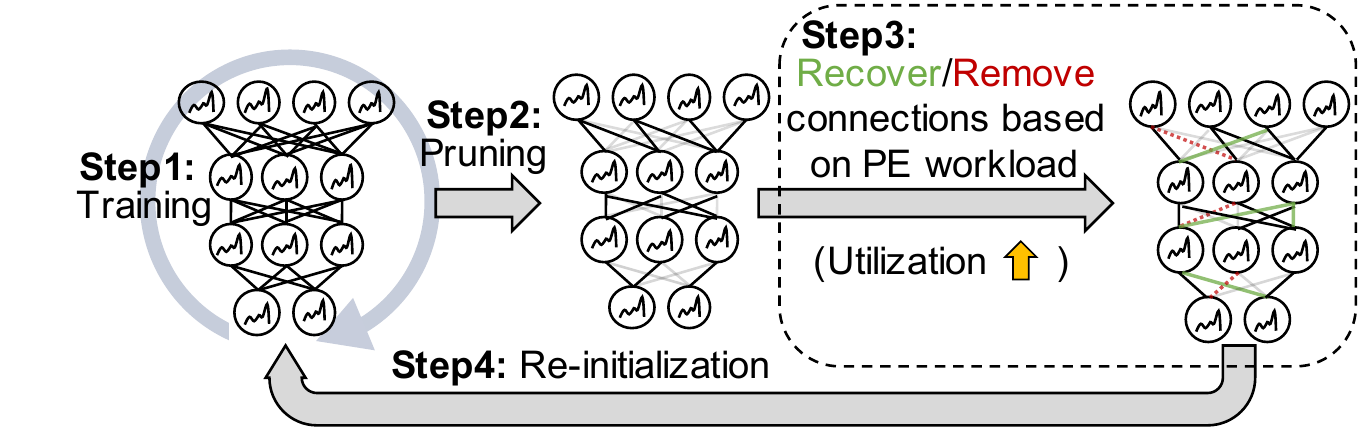} 
\\
\end{tabular}
\caption{ Illustration of the concept of the proposed u-Ticket. Our u-Ticket consists of training (\textbf{step1}), pruning (\textbf{step2}), adjusting weight connections based on workload (\textbf{step3}), and re-initialization (\textbf{step4}). We repeat these steps for multiple rounds.
Please note that the standard LTH method consists of training (\textbf{step1}), pruning (\textbf{step2}), and re-initialization (\textbf{step4}), which does not consider the utilization of the pruned SNNs.}

\label{fig:method:lth_vs_uticket}
\end{center}
\end{figure}

To this end, we propose u-Ticket, an iterative workload-balanced pruning method for SNNs that can effectively achieve high sparsity of weights and minimize the workload imbalance problem. 
Our method is based on Lottery Ticket Hypothesis (LTH) \cite{frankle2018lottery}, which states that sub-networks with similar accuracy can be found in over-parameterized networks by repeating \textit{training-pruning-initialization} stages.
Different from the standard LTH method \cite{kim2022exploring} where the pruned networks are naively used for the next round, we either remove or recover weight connections to balance workloads across all PEs before sending the networks to re-initialization (see Fig. \ref{fig:method:lth_vs_uticket}).

Compared to prior workload-balancing methods (see Fig. \ref{intro_comp}), the u-Ticket approach improves PE utilization by up to 100\% (70\% for \cite{gospa} and 92\% for \cite{sparten}) while maintaining filter sparsity of 98\% (60\% for \cite{gospa} and 35\% for \cite{sparten}), at iso-accuracy with the standard LTH-based pruning baseline \cite{kim2022exploring}. Furthermore, 
since our method balances the workload during the pruning process, u-Ticket does not incur additional hardware overhead for deployment.

We summarize the key contributions as follows:
\begin{enumerate}

\item  We propose u-Ticket, which discovers highly sparse SNNs with optimal PE utilization.
The discovered sparse SNN model achieves a similar level of accuracy, weight sparsity, and spike sparsity with the standard  LTH baseline \cite{kim2022exploring} while improving the utilization up to $100\%$. 

\item By balancing the workload, u-Ticket reduces the running latency and energy cost by up to $76.9\%$ and $63.8\%$, respectively, compared to the standard LTH method. 
\item We extend the prior sparse accelerator\cite{gospa} and propose an energy estimation model for sparse SNNs.

\item  To validate the proposed u-Ticket, we conduct experiments on two representative deep architectures (i.e., VGG-16~\cite{vgg} and ResNet-19~\cite{resnet}) across four public datasets including CIFAR10~\cite{krizhevsky2009learning}, Fashion-MNIST~\cite{xiao2017fashion}, SVHN~\cite{netzer2011reading}, and CIFAR100~\cite{krizhevsky2009learning}.
\end{enumerate}

\begin{figure}[h]
\centerline{\includegraphics[width=0.99\linewidth]{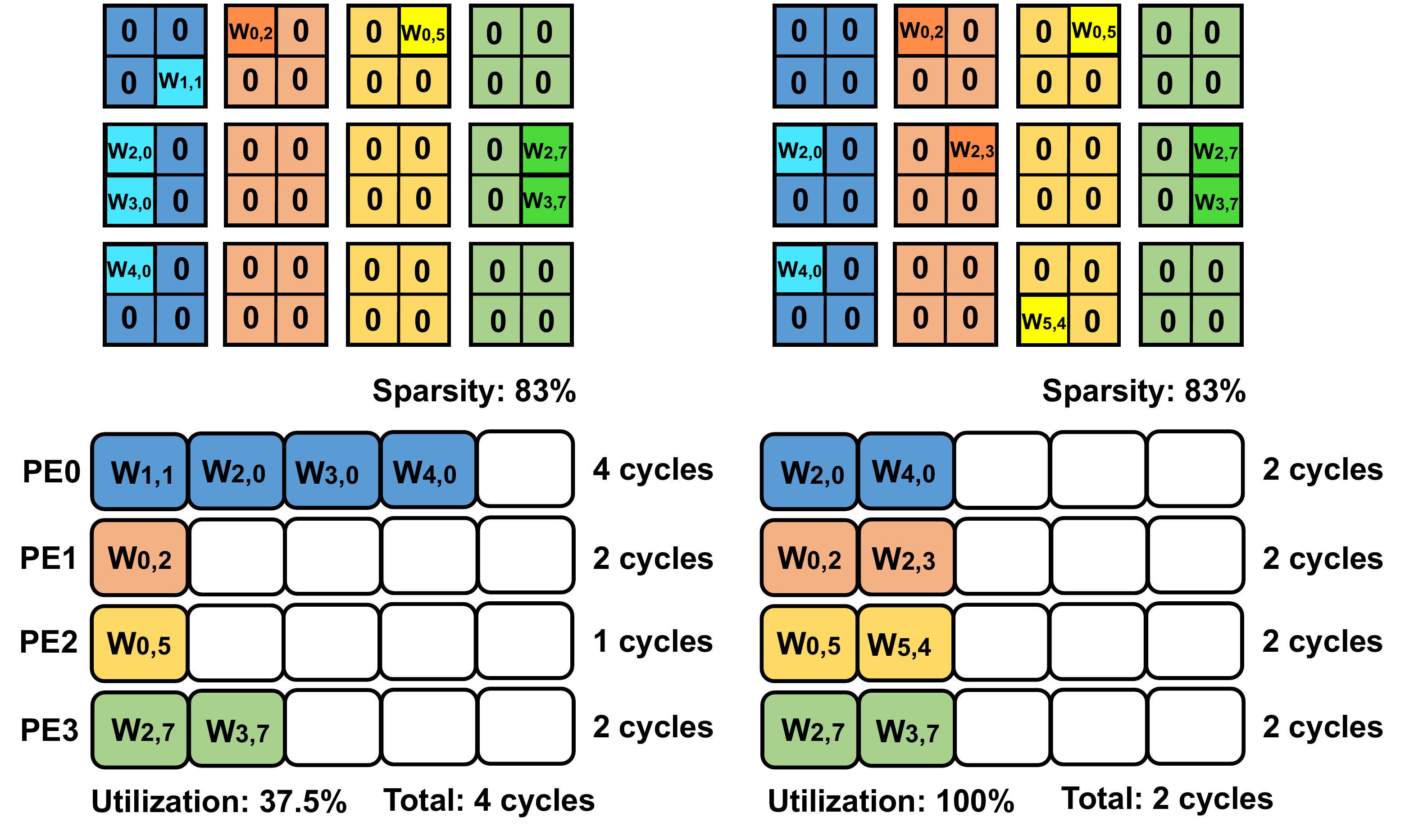}}
\caption{Example utilization and latency resulted from imbalance and balanced workload under the same model sparsity. With the unstructured pruning, non-zero weights will have a random distribution across four groups, thus leading to unbalanced workloads across PEs, as shown on the left side (PE0 has four weights assigned, while PE1 and PE2 only have one).}
\vspace*{-1em}
\label{imbalance_u}
\end{figure}

\section{Related Works}
\subsection{Pruning for Spiking Neural Networks}
Recently, there has been a significant growing interest in exploring spiking neural networks (SNNs) as the new generation of low-power deep neural networks under the context of edge machine learning~\cite{spike_nature,roadmap_snn}. One of the main groups of SNN research focuses on the compression of the size of the networks, which is very important on edge deployment, where the memory resources are usually very limited. This work mainly focuses on pruning, one of the most popular network compression techniques.

Pruning is a widely studied technique in neural network compression that aims to reduce the size of a neural network by removing unnecessary connections or weights while maintaining its accuracy. Researchers have extensively studied the pruning for SNNs to meet the limited memory resources on the edge~\cite{rathi2018stdp,neftci2016stochastic,deng2021comprehensive,guo2020unsupervised,chen2021pruning,shi2019soft,han2022adaptive,kim2022exploring}.

In \cite{rathi2018stdp,neftci2016stochastic}, the weight connections are removed if their magnitudes are below a pre-set threshold. In~\cite{chen2021pruning}, the weight connections are removed according to the magnitude of the gradients. In~\cite{deng2021comprehensive}, the ADMM optimization method is adopted to prune the SNNs. Both~\cite{guo2020unsupervised} and~\cite{shi2019soft} propose adaptive weight pruning for SNNs. The unsupervised online weight pruning is proposed by~\cite{guo2020unsupervised} for SNNs, while~\cite{shi2019soft} proposes the supervised training for the pruning masks for SNNs. A more bio-plausible pruning method for SNNs is recently proposed by \cite{han2022adaptive}, where the dendritic spine plasticity-based synaptic constraints are incorporated during the pruning process. 

However, all those prior pruning works for SNNs are limited to relatively shallow networks ($<$ 10 layers). In~\cite{kim2022exploring}, a lottery ticket hypothesis (LTH) based pruning method~\cite{frankle2018lottery} is proposed for SNNs that can efficiently prune over $95\%$ of the weight connections on very deep spiking neural networks (e.g., ResNet-19\cite{resnet}). Since the~\cite{kim2022exploring} shows the best performance on the deep SNN architectures (e.g., VGG-16 and ResNet-19) that many state-of-the-art SNN works adopt, we study our utilization recovery method for the LTH-based SNNs \cite{kim2022exploring}. 

Our approach differs from other LTH-related studies.
Because LTH introduces a high compression ratio with minimal performance degradation, researchers have explored its implications from various angles. Applying LTH beyond image classification is one of the primary directions, such as in natural language processing \cite{chen2020lottery}, graph neural networks \cite{chen2021unified}, and object detection \cite{girish2021lottery}. Another line of work also proves the theoretical background of LTH, suggesting that the identified initial parameters might be strongly tied to the identified sparse structure \cite{malach2020proving,pensia2020optimal,burkholz2022most,da2021proving}. The other recent work~\cite{yao2023probabilistic} studies the relationship between the LTH pruning and the LIFs' firing probability. It further theoretically proves that the LTH holds in SNNs. Our work, different from previous research, focuses on improving the hardware efficiency (i.e., the workload imbalance problem) of the LTH-pruned SNNs.

Please note that the SNNs we focus on in this work are the ones that are trained with backpropagation through time (BPTT). This group of SNNs performs superior accuracy in many complex vision tasks~\cite{kim2022exploring,spike-bp,direct_snn,tdbn}. Other groups of SNNs exist that are trained differently. For example, SNNs trained with spike-timing-dependent plasticity (STDP)~\cite{stdp_article_1, vaila2020deep, rathi2018stdp} and SNNs converted from pre-trained artificial neural networks~\cite{conversion_1,conversion_2,conversion_3}. Also, many other compression methods have been proposed to reduce the on-chip resource requirements for SNNs~\cite{lee2024tt,yin2023mint}. Those optimizations are orthogonal to the u-Ticket method.

\subsection{Utilization Recovery Methods}
In~\cite{gospa} and~\cite{eie}, deep FIFO queues build up a backlog of workloads and thus alleviate workload imbalance passively. On the other hand,~\cite{sparten} and~\cite{column} address the workload imbalance problem more systematically. In~\cite{sparten}, an offline shuffle of the weight positions is done to balance the workload in finer granularity. Similarly,~\cite{column} introduces a new training method to pack the sparse weights into a denser group for improving workload utilization. They both introduce additional hardware for permuting and unshuffling back the partial sum to the appropriate position. Authors in~\cite{ese} propose a utilization-aware pruning method for ANNs on speech recognition tasks. However, such a pruning method has not been explored in image classification. Moreover, they require the FIFO queues to achieve decent utilization ($\sim $90\%). 
Unlike prior works, our method recovers the utilization of SNNs in image classification tasks without requiring additional hardware units for implementation.
Also, we achieve significantly higher utilization ($\sim $100\%) compared to the previous works.
Table~\ref{tb:1} summarizes the comparison between our and other utilization recovery methods to solve the workload imbalance problem.

\begin{table}[t]

\centering
\caption{Comparison with prior works on workload imbalance problem. HW denotes hardware.}
\vspace*{-1mm}
\begin{adjustbox}{max width =\linewidth}
\begin{tabular}{@{\extracolsep{3pt}}lccc}

\toprule   
 \textbf{Accelerator} &  \textbf{Weight Sparsity$^1$} & \textbf{Extra HW} & \textbf{Utilization$^2$}\\
\midrule
% \addlinespace[0.4em]
{EIE}\cite{eie}   & 77\% & FIFOs & $\sim $75\%\\ 
\addlinespace[0.4em]
{GoSPA}\cite{gospa}   & 62\% & FIFOs & $\sim $70\%\\
\addlinespace[0.4em]
{SparTen}\cite{sparten}   & {35\%} & Shuffle Units & $\sim $95\%\\
\addlinespace[0.4em]
{Column}\cite{column}   & {84\%} & Permute Units & $\sim $90\%\\
% \addlinespace[0.4em]
\midrule
\textbf{u-Ticket (ours)}  & {98\%} & Not Required & $\sim $100\%\\
\bottomrule
\multicolumn{4}{l}{$^1$Measured on the VGG-like network reported in the prior works.}\\
\multicolumn{4}{l}{$^2$We report the median utilization from the prior works.}
\end{tabular}
\label{tb:1}
\end{adjustbox}
\end{table}

\section{Background}

\subsection{Spiking Neural Networks}

Spiking Neural Networks (SNNs) process the unary temporal signal through multi-layer weight connections.
Instead of a ReLU neuron for a non-linear activation, recent SNN works use a Leaky-Integrate-and-Fire (LIF) neuron, which contains a memory called membrane potential.
The membrane potential captures the temporal spike information by storing incoming spikes and generating output spikes accordingly.
Suppose a LIF neuron $i$ has a membrane potential $u_{i}^{t}$ at timestep $t$. We can formulate the discrete neuronal dynamics \cite{wu2018spatio,fang2021incorporating} by:
\begin{equation}
    u_i^t = \lambda  u_i^{t-1} + \sum_j w_{ij}s^t_j.
    \label{eq:LIF}
\end{equation}
Here, $\lambda$ is the leaky factor for decaying the membrane potential through time. The $s^t_j$ stands for the output spike from a neuron $j$ at timestep $t$.
The $w_{ij}$ denotes a weight connection between neuron $j$ in the previous layer and neuron $i$ in the current layer.
If the membrane potential reaches a firing threshold, the neuron generates an output spike, and the membrane potential is reset to zero.
Like ANNs, we train the weight connection $w_{ij}$ in all layers.
Our weight optimization is based on the recently proposed surrogate gradient learning, which assumes an approximated gradient function for the non-differentiable LIF neuron \cite{neftci2019surrogate}.
We use $tanh(\cdot)$ approximation following the previous work \cite{fang2021incorporating}.

\subsection{Lottery Ticket Hypothesis}

Lottery Ticket Hypothesis (LTH) \cite{frankle2018lottery} has been proposed where they found a dense neural network contains sparse sub-networks (\ie winning tickets) with similar accuracy compared to the original dense network.
The winning tickets are found by multiple rounds of magnitude pruning operations. Specifically, suppose we have a dense network $f(x;\theta)$ with randomly initialized parameter weights $\theta \in \mathbb{R}^{n}$.  
In the first round, the dense network $f(x;\theta)$ is trained to convergence (\textbf{step1} in Fig. \ref{fig:method:lth_vs_uticket}). Based on the trained weights, we prune $p\%$ weight connections with the lowest absolute weight values (\textbf{step2} in Fig. \ref{fig:method:lth_vs_uticket}).
We represent this pruning operation as a binary mask $m \in \{0, 1\}^{n}$.
In the next round, we re-initialize the pruned network with the original initialization parameters $f(x;\theta \odot m)$ (\textbf{step4} in Fig. \ref{fig:method:lth_vs_uticket}), where $\odot$ represents the element-wise product.
The \textit{training-pruning-initialization} stages are repeated for multiple rounds.
In the SNN domain, Kim~\etal~\cite{kim2022exploring} recently applied LTH to deep SNNs, resulting in high weight sparsity ($\sim $98\%) for VGG and ResNet architectures.
However, they do not consider the workload imbalance problem in sparse SNNs.
Different from the previous work, we adjust weight connections for improving utilization at each pruning round \textbf{step3}, which reduces up to $77\%$ latency and $64\%$ energy cost compared to the standard LTH \cite{kim2022exploring} while maintaining both sparsity and accuracy.

\subsection{Workload Imbalance Problem}
\label{problem_sec}
In the context of neural network accelerators, dataflow refers to the hardware's input and weight mapping strategy.
To this effect, recent works \cite{scnn,sparten,gospa,spinalflow,sata} have demonstrated the efficacy of the weight stationary dataflow towards efficient deployment of sparse networks and SNNs. 

For weight-stationary dataflow, weights are cast to different PEs and stay inside the PE until they are maximally reused across all the relevant computations. More specifically, during the running time, depending on the memory capacity of the hardware, each layer's filter kernels will be grouped in a chosen pattern and sent to each PE. As shown in Fig. \ref{imbalance_u}, the number of non-zero elements (or workload) allocated to each PE varies significantly due to the randomness in unstructured pruning. Moreover, the workload imbalance is persistent irrespective of the grouping method chosen. Note that here, we define the number of non-zero weights assigned to a PE as the workload.

In this case, the wasted resources in PEs are based on the difference between the largest workload and the average of all other workloads. To quantitatively measure the portion of non-wasted resources, we use the utilization metric\cite{loadimbalance}, given by 

\begin{equation}
\label{eq:imbalance}
    \mu = 1- 
    \frac{T_{max} - T_{avg}}{T_{max}} \cdot \frac{n}{n-1},
\end{equation}
$T_{max}$ and $T_{avg}$ are the longest and the average processing time among the PEs. $n$ is the number of PEs. The metric quantifies the percentage of processing time that the rest of the PEs, excluding the slowest ones, are engaging in useful work. 

\begin{figure}[t]
\centerline{\includegraphics[width=0.95\linewidth]{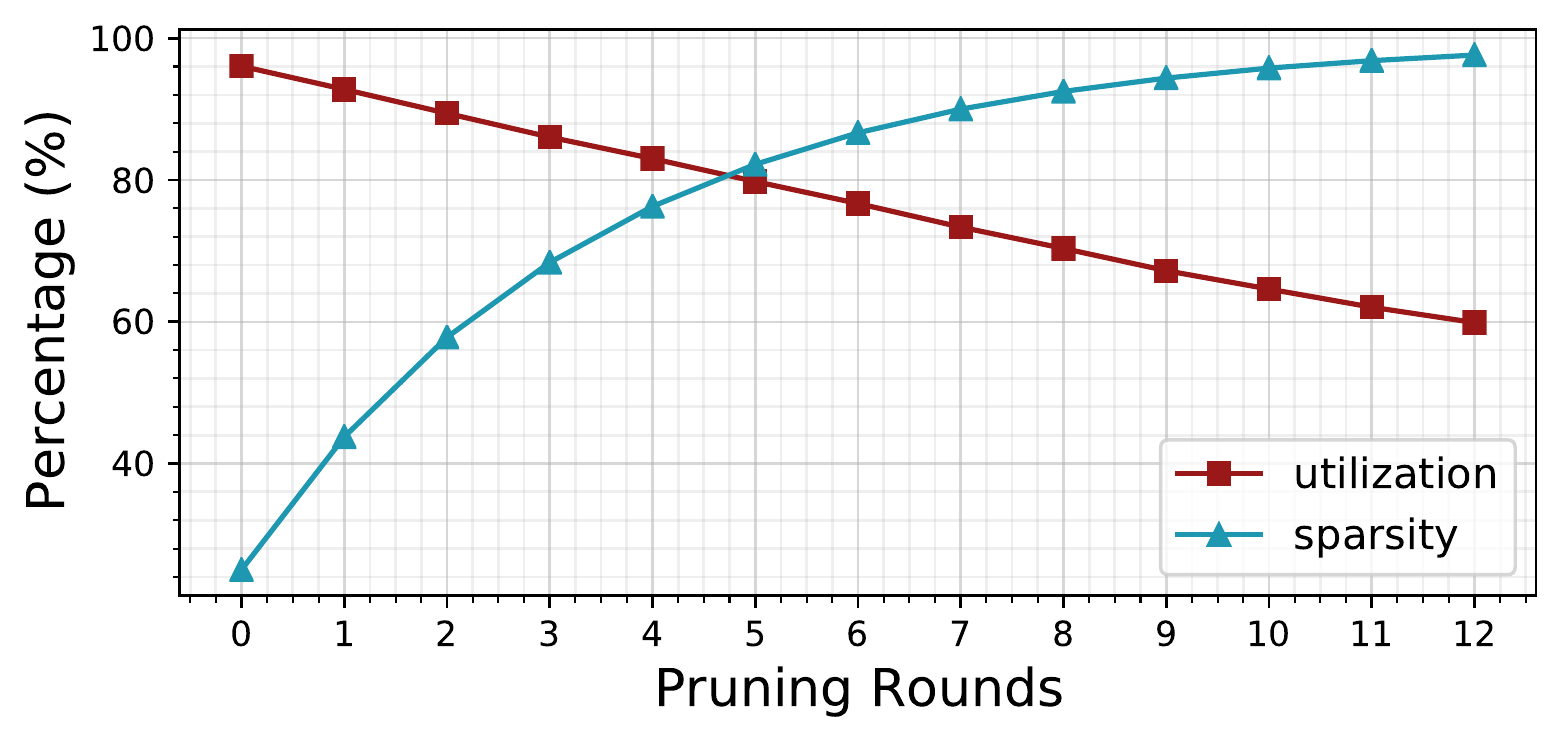}}

\caption{Sparsity and utilization across pruning rounds for the standard LTH method without utilization awareness. The pruning is done for 13 rounds on VGG-16 being trained for image classification on CIFAR10 with 16 PEs.} 

\label{lth_problem}
\end{figure}

In Fig. \ref{lth_problem}, we show how the utilization degrades as the weight sparsity of the SNN increases in the standard LTH method \cite{kim2022exploring}. The preliminary result shows that in the final round, the utilization can be as low as $59\%$ on VGG-16 CIFAR10. Here, we assume that the total number of PEs is 16, and the utilization is averaged across all layers (weighted by parameter count).

\section{u-Ticket}

To resolve the workload imbalance problem, we propose u-Ticket, where we achieve high utilization in sparse SNNs during iterative pruning. 
In this section, we first present the algorithm to train sparse SNNs while maintaining high utilization.
We then provide details of the proposed PE design and the energy model to map the u-Ticket model on the hardware.

\subsection{Algorithmic Approach}
\subsubsection{\textbf{Algorithm Overview}}

Our u-Ticket pruning consists of multiple rounds similar to LTH \cite{frankle2018lottery}.
For each round, we train the networks till convergence, prune the low-magnitude weight connections, balance the workload of PEs by recovering or removing the weight connections, and finally re-initialize the weights.

The main idea is to ensure a balanced workload between PEs after each unstructured pruning round.

The overall u-Ticket process is described in Algorithm \ref{alg:u_rec}.
For each round, the pruned SNN from the previous round is re-initialized. After that, the model is trained and pruned where we obtain connectivity mask $\hat{m}_i$ with imbalanced PE workloads.
To increase the utilization, we first compute the workload for each PE, constructing the PE workload list $W^l$ for each layer. Based on the $W^l$, we calculate the average workload $w_{avg}^l$ for layer $l$.
Then, we go through each workload $w$ in $W^l$ and randomly recover ($w_{avg}-w$) of weight connections if the PE's workload $w$ is smaller than the average workload $w_{avg}^l$.
Otherwise, the number of weight connections is pruned by ($w-w_{avg}$).
After the workload adjustment, every workload $w$ will have the same magnitude to ensure the optimal utilization $\mu$.
We repeat the stages above for $N$ rounds.

\subsubsection{\textbf{Design Choice of Workload Balance}}
There are three main design metrics to be considered for our workload balancing process: workload mapping granularity, workload checking granularity, and the workload balancing method.

\noindent \textbf{Workload mapping granularity.} \indent In our u-Ticket, we assume the following procedure of mapping the weights into the PEs. For each PE, we will assign it with all the non-zero weights in one filter. Those non-zero weights will stay stationary inside the PE to fully utilize the weight-reuse across all timesteps. This weight mapping method is adopted by many recent SNN accelerator designs~\cite{spinalflow,sata,ptb}. Moreover, similar weight-stationary mapping is also adopted in many recent sparse accelerator designs due to its dataflow efficiency under the context of sparse neural networks~\cite{gospa,scnn,sparten}.

\noindent \textbf{Workload checking granularity.} \indent In our method, we use average workload $w_{avg}^l$ across all PEs at layer $l$ as the reference to recover/remove weight connections.
The reason behind such a design choice is as follows: 

\noindent(1) If we look at only partial PE workloads to decide on a reference workload, though it will reduce the complexity of getting the average workload, it will inevitably bring a sub-optimal solution~\cite{frankle2018lottery}.

\noindent(2) The cost of checking all PE workloads to get the average workload is negligible compared to the overall iterative training-pruning-initialization process. We find that on an RTX 2080Ti GPU, the total time cost of traversing through all PEs to get the average workloads is only 0.2\% of one complete LTH searching round. 

\noindent \textbf{Workload balancing method.} \indent In our workload balancing method, we randomly recover and remove the weights to get the optimal workload. There are other criteria to choose the weights. For example, the magnitude of the weights is a very common option~\cite{lss,han2015deep}.

\begin{algorithm}[t]\small
        \caption{u-Ticket}\label{alg:u_rec}
       \textbf{Input}: SNNs $f(x;\theta)$ with randomly-initialized parameter weights $\theta \in \mathbb{R}^{n}$, connectivity mask $m_i \in \{0, 1\}^{n}$ at iteration $i$, total pruning round $N_{Round}$, total number of layer $L$, number of PEs $n$, Workload of a PE $d$, Workload list of a layer $W^{l}$.
       \\
      \textbf{Output}:  Pruned  $f(x;\theta_{trained} \odot m_N)$ 
      \begin{algorithmic}[1]
        % \State{\textbf{begin}}
        %
        \State {initialize $m_1$ with $1$} \vspace{0.25cm}

        \For{$i \gets $ 1 to $N_{Round}$}\vspace{0.1cm}
        \State {$f(x;\theta \odot m_i)$} \vspace{0.1cm}   
        
        \State {$f(x;\theta_{trained} \odot m_i) \gets  Train(f(x;\theta \odot m_i))$} \vspace{0.1cm}
        % \Comment{Train the pruned SNNs}
         \State {$\hat{m}_{i} \gets Prune (f(x;\theta_{trained} \odot m_i)$)} \vspace{0.1cm}
 
        \For{$l \gets $ 1 to $L$}  \vspace{0.1cm}

            \State{$W^{l} \gets \texttt{GetWorkload}(f(x;\theta^{l}_{trained} \odot \hat{m}^{l}_{i}), n$)}\vspace{0.1cm}
            \State{$d^{l}_{avg} \gets \texttt{GetAverage}(W^{l})$}\vspace{0.1cm}
                \For{$d$ in $W_{l}$}\vspace{0.1cm}

                \If{$d < d^{l}_{avg}$}
                    \State {$m^{l}_{i+1} \gets  \texttt{Recover}(\hat{m}^{l}_{i},  d_{avg}-d)$}
                \Else
                    \State {$m^{l}_{i+1} \gets  \texttt{Remove}(\hat{m}^{l}_{i},  d-d_{avg})$}\vspace{0.1cm}
                \EndIf
                \EndFor
        \EndFor
    \EndFor
      \end{algorithmic}
      
          \label{algorithm: overall}
\end{algorithm}

\begin{figure}[t]
\centerline{\includegraphics[width=\linewidth]{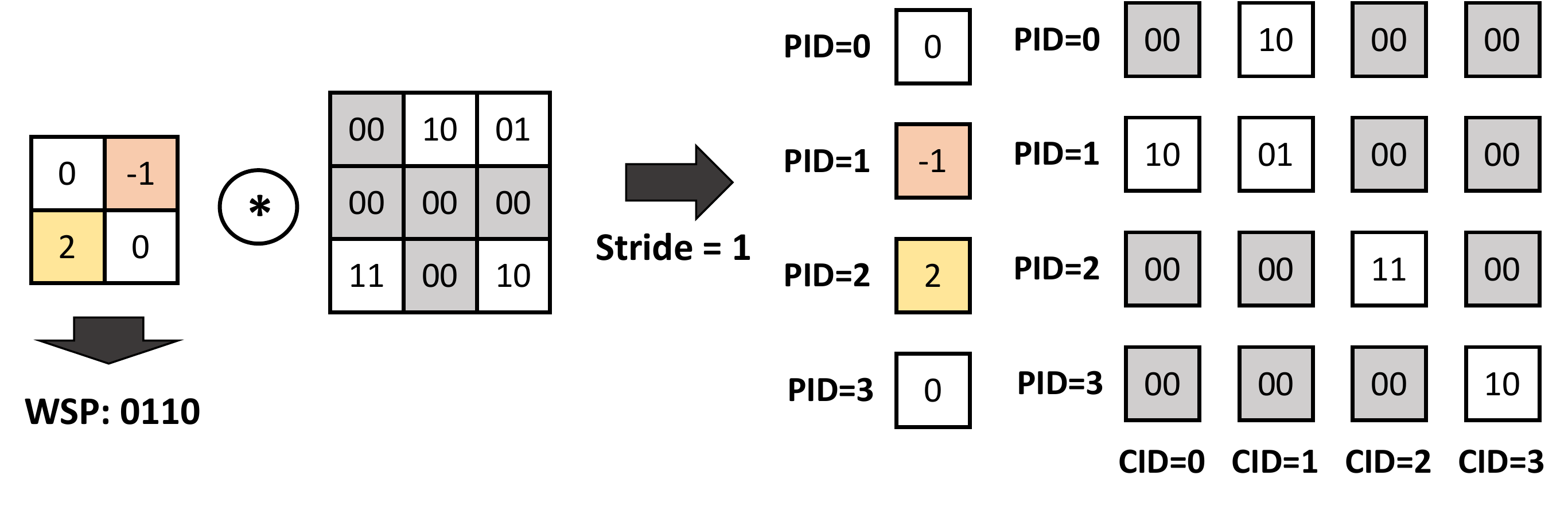}}

\caption{Illustration of the weight sparsity pattern (WSP), the position ID (PID), and the (convolution ID) CID.} 

\label{fig:index}
\end{figure}

\begin{figure*}[t]

\centerline{\includegraphics[width=0.8\linewidth]{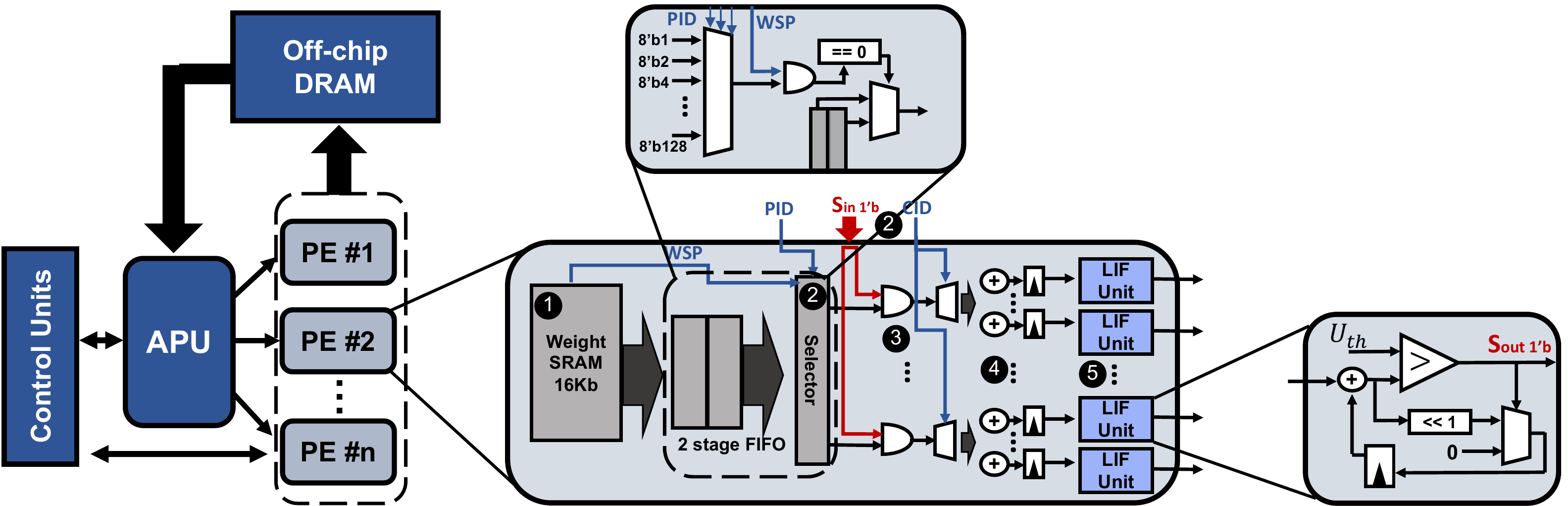}}
\caption{Overall architecture and the detailed inner architecture of PE. Here, APU denotes the activation processing unit.}
\label{arch}

\end{figure*}

We empirically find that randomly choosing the weights to be recovered and removed has very similar accuracy to the criteria-based choosing method. Meanwhile, the random-based choosing method has a better searching complexity with $O(n)$, while the criteria-based choosing method has at least a complexity of $O(n\cdot\log n)$.

\subsection{Hardware Mapping}

\subsubsection{Processing Elements (PEs)}

We must map the sparse SNN to a proper hardware design to get an accurate energy estimation. We develop our PE design based on \cite{gospa}, one of the state-of-the-art sparse accelerators, to support running sparse SNNs. Please note that our method of balancing the workloads works on any sparse accelerator design as long as it utilizes the weight stationary dataflow.

First, the non-zero weights, input spikes, and their corresponding metadata (index) are read from the DRAM. The weights are represented in weight sparsity pattern (WSP) \cite{gospa}, while the spike activations are represented in standard compressed sparse row (CSR) format. We use four timesteps for the SNN in our experiments, thus we can group every two activations into one byte (each activation has four unary spikes.)

Then, an activation processing unit (APU, outside PEs) filters out the zero activation (0-spikes across four timesteps) and sends the non-zero activation together with their position indices (decoded from CSR) to the PE arrays. The position indices help to match the non-zero weights and activation in 2-D convolution. 

In Fig.~\ref{fig:index}, we further illustrate the position indices used in this work. We use PID (Position ID) and CID (Convolution ID) to match the valid combination of spike activations and weights (both non-zero). We explain the assignment of PID and CID on the right of Fig.\ref{fig:index}. For an unrolled 2-D convolution map, all the activations involved with the same inner product (in the same column) share the same CID. The activations on the same row share a PID, as they correspond to the same weight, which is also assigned the same PID value. With this CID/PID matching, it is very convenient to match the non-zero pairs of weights and spike activations~\cite{gospa}.

Each PE contains four 16-bit AND gates, 256 24-bit accumulators, and one 1024 $\times$ 16 bits SRAM-based scratch-pad. We extend the 256 accumulators with 256 LIF units to generate the output spikes. Each LIF unit equips four 24-bit registers for storing the membrane potential across four timesteps.

Fig. \ref{arch} illustrates the overall architecture and the computation flow inside the PE. We process the network in a tick-batched manner \cite{spinalflow}. At step \circled{1}, the non-zero weights and their WSPs are mapped to each PE. At step \circled{2}, the spike activation $S_{in}$ and their position indices are sent to PE. The selector unit will output the matched non-zero weight based on the weight's WSP and the activation's position index. At step \circled{3} and \circled{4}, the dot-product operations between the input spike and the matched weights are carried out, and the partial sums are stored according to their position index. At step \circled{5}, the partial sums for each time step are sequentially sent to the LIF units to generate the output spikes for each time step. Note that steps \circled{3} - \circled{5} need to be repeated four times to match the four timesteps used in our SNN model (only 1 bit of $S_{in}$ is cast to the PE at a time in step \circled{2}).

\subsubsection{Energy Modeling}
\label{sec:model}
We do the simulation for the full architecture. Since u-Ticket balances the workloads between PEs, most improvements can be found at the PE level. Thus, this work focuses on energy estimation at the PE level.
We extend the energy model from \cite{sata} to estimate the total energy:
\begin{equation}\label{eq:pe_energy_total}
E_{total}= N_{work} \cdot (E_{PE}^{d}\cdot(1-S_{in}^{spa}) + E_{PE}^{l}) + N_{idle} \cdot E_{PE}^{l},
\end{equation}
where $E_{PE}^{d}$ and $E_{PE}^{l}$ are the dynamic and leakage energy of a single PE processing one input spike. As shown in \cite{sata}, there is no extra cost for skipping the zero-spike computation in SNNs. In this work, we directly apply the input spikes as the enable signal of the accumulators and LIF units. In this way, we can stop those circuits from flipping when there are incoming zero spikes. Thus, we directly apply the term of spike sparsity, $S_{in}^{spa}$, in Eqn. \ref{eq:pe_energy_total} to approximate the dynamic energy saving by skipping the zero spikes. Here $N_{work}$ is defined as the total work cycles in which PEs are doing useful work and $N_{idle}$ denotes the total cycles in which PEs are waiting in an idle state.

\section{EXPERIMENT}
\subsection{Experimental Settings}
\subsubsection{Software Configuration}
First, to validate the u-Ticket pruning method, we evaluate our u-Ticket methods on four public datasets: CIFAR10~\cite{krizhevsky2009learning}, Fashion-MNIST~\cite{xiao2017fashion}, SVHN~\cite{netzer2011reading}, and CIFAR100~\cite{krizhevsky2009learning}. We choose two representative deep network architectures: VGG-16 \cite{vgg} and ResNet-19 \cite{resnet}. We implement the networks on PyTorch and set the timesteps $T$ to 4 for all experiments. We use the state-of-the-art direct encoding technique that has been shown to train SNNs on image classification datasets with very few timesteps. We use the same training configurations in Table.~\ref{tab:hyper}.

\begin{table}[t]
\centering
\renewcommand*{\arraystretch}{1.3}
\caption{SNN training hyperparameters for our u-Ticket method.}
\label{tab:hyper}
\vspace{-2mm}
\begin{adjustbox}{max width =0.8\linewidth}
\begin{tabular}{lcc}
\toprule
Parameters & Description & Quantity \\ \midrule
Batch Size & - & 128 \\
Optimizer & - & SGD \\
$T$ & timesteps & 4 \\
$\gamma$ & learning rate & 1e-1 \\
$\lambda$ & weight decay & 5e-4 \\
$\mu$ & momentum & 0.9 \\
$\tau$ & membrane potential leak & 0.75 \\
$v_{th}$ & firing threshold & 1 \\
reset mode & - & hard \\
epoch & number of training epochs & 150 \\ \bottomrule
\end{tabular}
\end{adjustbox}
\end{table}

\subsubsection{Hardware Configuration}
 We report the utilization, latency, work cycles, and idle cycles based on our PyTorch-based simulator, which simulates the running-time distribution of the weights to PEs. We distribute the weights from one filter into 16 PEs. The PE level energy is estimated with the model in Section~\ref{sec:model} with all computing units synthesized in Synopsys Design Compiler at 400MHz using 32nm CMOS technology and the memory units simulated in CACTI. We set the standard LTH method \cite{kim2022exploring} without utilization-awareness as our baseline and use the same estimation model to get the speed-up and energy results.

\begin{table}[t!]

\centering

\caption{Comparison of accuracy, filters' sparsity, and spikes' sparsity between our method and the standard LTH method.}
\vspace*{-1mm}
\begin{adjustbox}{max width =\linewidth}
\begin{tabular}{@{\extracolsep{4pt}}llccc}

\toprule   

 {Dataset} & {Method} & {Acc.(\%)} & {Sparsity(\%)} & {Sparsity(\%)}\\
 &  &  & {(filters)} & {(spikes)}\\
\midrule
\multicolumn{5}{c}{{VGG-16}\cite{vgg}} \\
\midrule

{CIFAR10}  & LTH\cite{kim2022exploring} & \textbf{91.0}& 98.2 & 84.8\\
         & {u-Ticket} (ours) & 90.7 & \textbf{98.4} & \textbf{85.9} \\
\addlinespace[0.4em]
{FMNIST}  & LTH\cite{kim2022exploring} & \textbf{94.6} & 98.2 & \textbf{83.9}\\ 
        & {u-Ticket} (ours) & 94.0 & \textbf{98.5} & 81.4\\
\addlinespace[0.4em]
{SVHN}& LTH\cite{kim2022exploring} & \textbf{95.5} & 98.2 & \textbf{84.9}\\ 
      & {u-Ticket} (ours) & 94.8 & \textbf{98.5} & 80.1\\ 
\addlinespace[0.4em]
{CIFAR100}& LTH\cite{kim2022exploring} & \textbf{63.9} & 98.2 & 81.9\\ 
      & {u-Ticket} (ours) & 63.1 & \textbf{98.2} & \textbf{82.0}\\ 
\midrule
\multicolumn{5}{c}{{ResNet-19}\cite{resnet}} \\
\midrule

{CIFAR10}  & LTH\cite{kim2022exploring} & \textbf{91.0} & 97.6 & 64.1\\ 
         & {u-Ticket} (ours) & 90.3 & \textbf{98.4} & \textbf{68.3}\\
\addlinespace[0.4em]
{FMNIST}  & LTH\cite{kim2022exploring} & \textbf{94.4} & 98.2 & 60.1\\ 
        & {u-Ticket} (ours) & 93.3 & \textbf{99.0} & \textbf{62.9}\\
\addlinespace[0.4em]
{SVHN}  & LTH\cite{kim2022exploring} & \textbf{95.1} & 97.6 & 63.6\\ 
      & {u-Ticket} (ours) & 94.6 & \textbf{98.6} & \textbf{68.2}\\
\addlinespace[0.4em]
{CIFAR100}& LTH\cite{kim2022exploring} & \textbf{66.7} & \textbf{98.2} & 77.5\\ 
      & {u-Ticket} (ours) & 66.3 & 98.1 & \textbf{77.6}\\ 
\bottomrule
\end{tabular}
\label{valid_tab}
\end{adjustbox}
\end{table}

\begin{figure*}[t]
\begin{center}
\def\arraystretch{0.5}
\begin{tabular}{@{}c@{}c@{}c@{}c}
\includegraphics[width=0.25\linewidth]{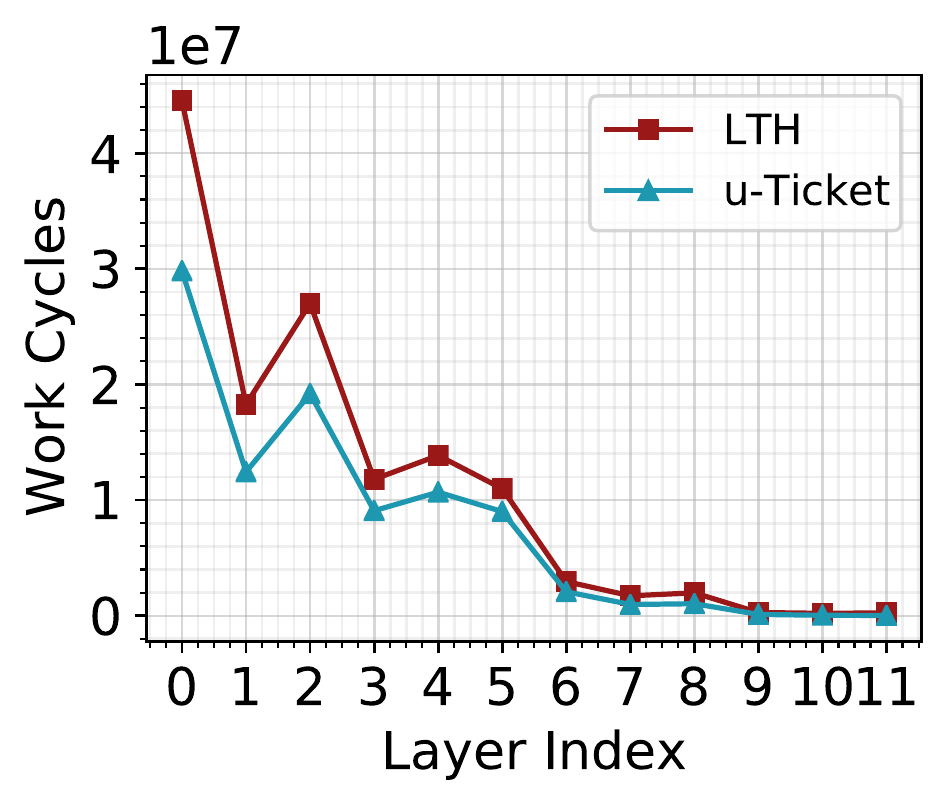} &
\includegraphics[width=0.25\linewidth]{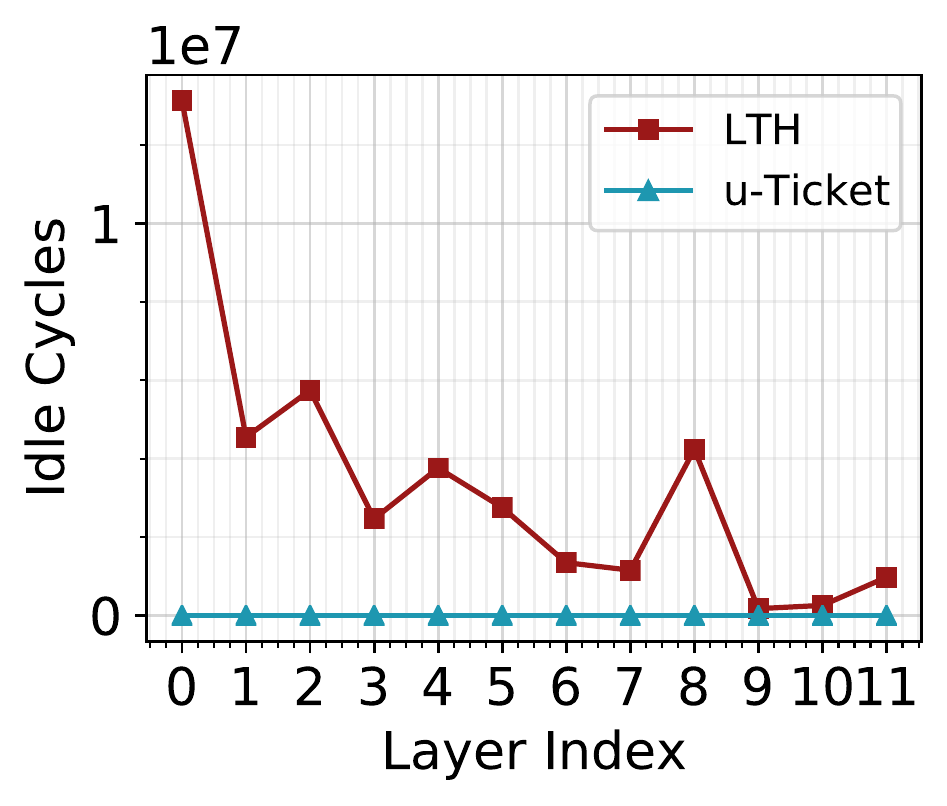}&
\includegraphics[width=0.25\linewidth]{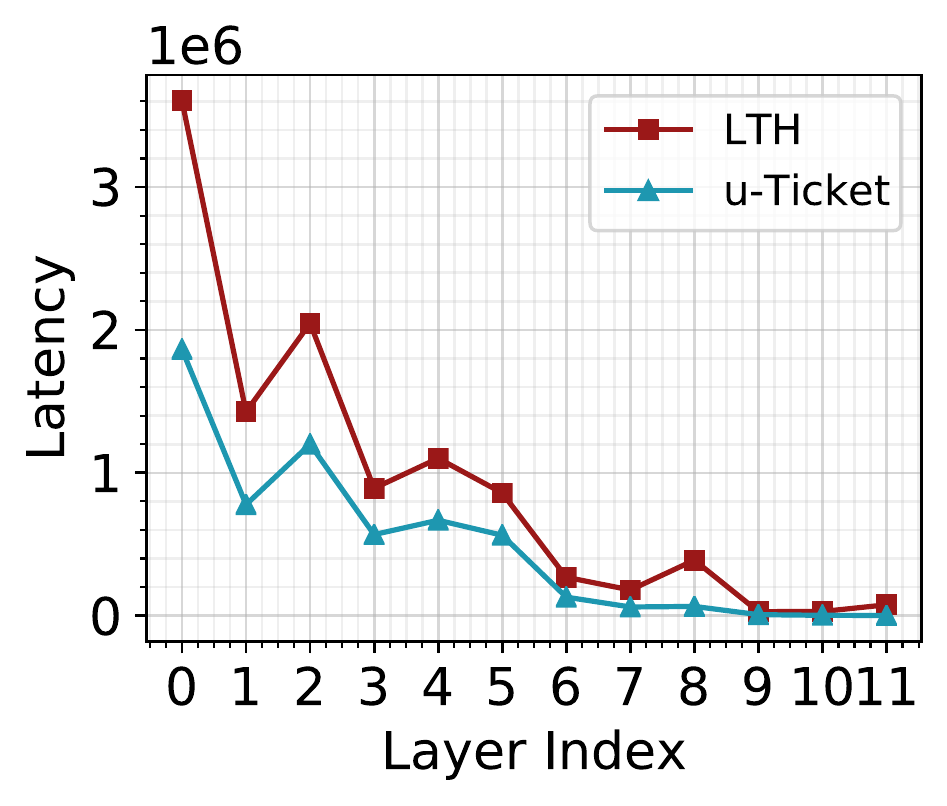} &
\includegraphics[width=0.265\linewidth]{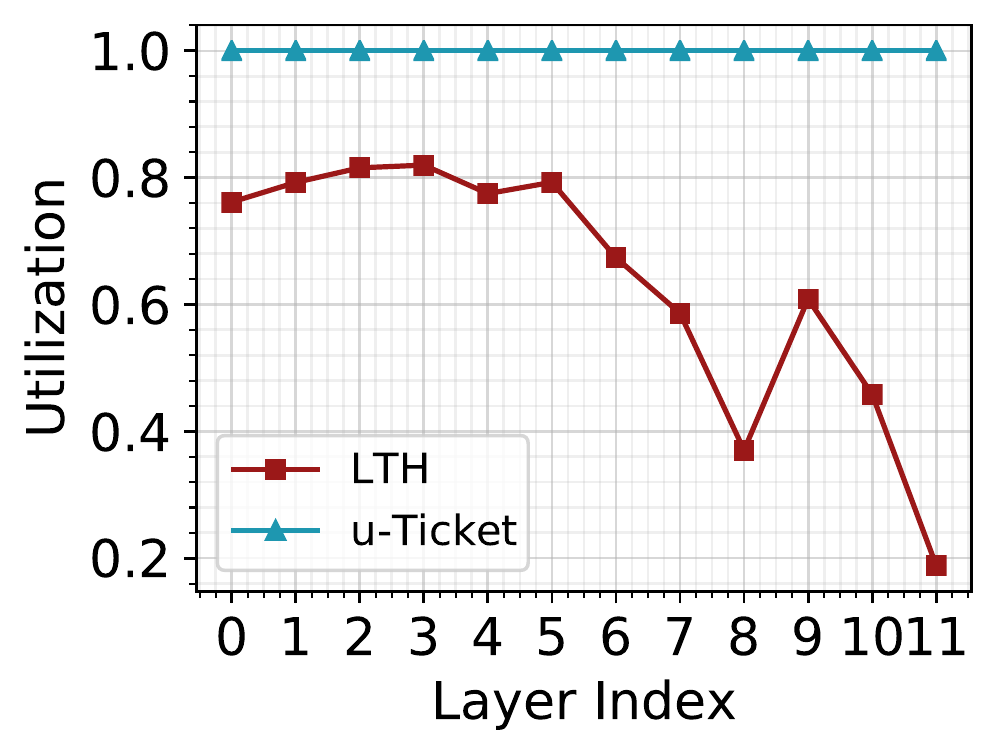} \\
{\hspace{3mm}(a)} & {\hspace{3mm}(b)} &{ \hspace{3mm}(c)}& { \hspace{5mm}(d) } \\
\end{tabular}
\vspace{-2mm}
\caption{The layerwise performance comparison between LTH and u-Ticket on four metrics, \ie  (a) work cycles, (b) idle cycles, (c) latency, (d) utilization. We conduct experiments with VGG-16 architecture on CIFAR10.
}
\label{fig:layerwise_speed}

\end{center}
\end{figure*}

\subsection{Experimental Results}
\subsubsection{Validation Result}
We summarize the validation results in Table \ref{valid_tab}. The results confirm that our method works well for deep SNNs (less than $\sim $1\% accuracy drop). We also compare the sparsity of filters and spikes between these two methods. u-Ticket has a slightly higher filter sparsity due to the extra reduction in weight connections to ensure balanced workloads for each PE. At the same time, u-Ticket keeps a similar level of spike sparsity on VGG-16 and has better spike sparsity on ResNet-19. While a higher spike sparsity will bring better energy efficiency, a spike sparsity that is too high will cause an accuracy drop in deep SNNs \cite{li2021differentiable}. This explains the accuracy-sparsity tradeoff on ResNet-19 (on average, 0.76\% accuracy drop with 3.5\% sparsity gain).

\begin{figure}[t]
\hspace{9mm}{\includegraphics[width=0.7\linewidth]{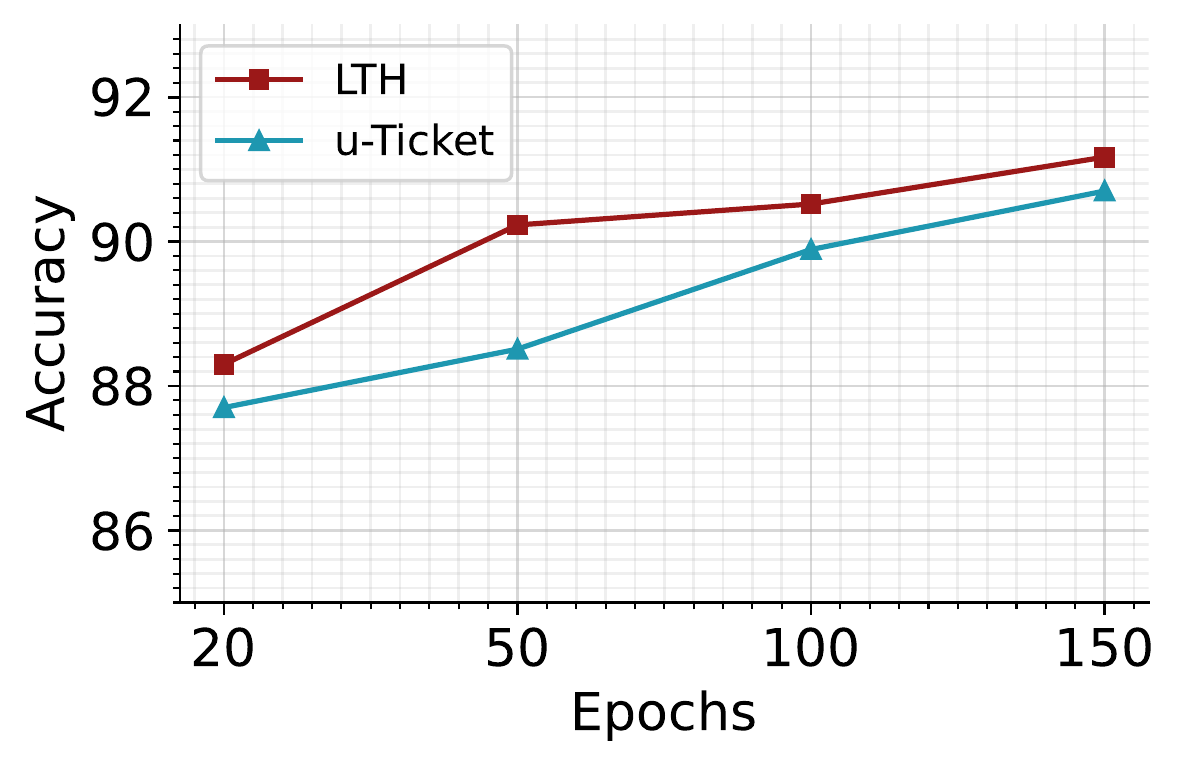}}
\vspace{-3mm}
\caption{Comparison of the convergence speed between the original LTH method and our method. The accuracy results are based on CIFAR-10 with VGG-16 networks.}
% \vspace*{-2em}
\label{fig:convergence}
\end{figure}

In addition, we show the convergence speed results between our method and the original LTH method in Fig.~\ref{fig:convergence}. The result shows that on the CIFAR-10 dataset with the VGG-16 network, our method only brings a slight convergence overhead in the first 50 epochs. After the 100 epoch, the convergence difference between the two methods is already negligible. The same trends can also be found on other datasets and networks.

\subsubsection{Hardware Performance}

We consider four metrics in this section (\ie work cycles, idle cycles, latency, and utilization). 

   \begin{itemize}
     \item \textbf{Work cycles} ($N_{work}$ in Eqn. 4): Sum of total work cycles for every PE across all the layers in the network. 
     \item \textbf{Idle cycles} ($N_{idle}$ in Eqn. 4): Sum of total idle cycles for every PE across all the layers in the network.
     \item \textbf{Latency}: Time required by PEs to process all the layers in the network. The latency is normalized with respect to the time required for a PE to process one input spike.
     \item \textbf{Utilization}: We use Eqn. \ref{eq:imbalance} to compute the utilization for each layer. 
     To compute the utilization of the network, we calculate the weighted average utilization. 
   \end{itemize}

The hardware improvement results are summarized in Table \ref{speed_tab}. By iteratively applying the utilization recovery during the pruning, u-Ticket can recover the utilization up to $100\%$ in the final pruning round, thus reducing almost all the idle cycles for PEs. Because of the re-balance of workloads among PEs, the network can leverage more parallelism from the PE array, thus significantly reducing the running latency. The number of work cycles stays similar on both networks. We further visualize the layerwise speedup results for VGG-16 on CIFAR10 in Fig. \ref{fig:layerwise_speed}. Overall, the layerwise work cycles and latency share a similar trend between the two methods. Furthermore, u-Ticket has more idle cycle reductions on earlier layers due to the larger feature map sizes. 

\begin{table}[t!]
\centering
\caption{Comparison of work cycles, idle cycles, latency, and utilization between u-Ticket and the standard LTH.}
\begin{adjustbox}{max width =0.9\linewidth}
\begin{tabular}{@{\extracolsep{4pt}}llcccc}
\toprule   

 {Dataset} & {Method} &{Work} & {Idle} & {Latency} & {Utilization} \\
 &  & \textbf{($\times 1e8$)} & \textbf{($\times 1e8$)} & \textbf{($\times 1e8$)} & \\
\midrule
\multicolumn{6}{c}{{VGG-16}\cite{vgg}} \\
\midrule

{CIFAR10}  & LTH\cite{kim2022exploring} & {1.34} & 0.41 & 0.11 & 0.59\\
         & {u-Ticket} (ours) & \textbf{0.94}  & \textbf{0.00} & \textbf{0.06} & \textbf{1.00} \\
\addlinespace[0.4em]
{FMNIST} & LTH\cite{kim2022exploring} & 1.10 & 0.42 & 0.10 & 0.57\\ 
        & u-Ticket (ours)& \textbf{0.81} & \textbf{0.00} & \textbf{0.05} & \textbf{1.00}\\
\addlinespace[0.4em]
{SVHN} & LTH\cite{kim2022exploring} & 1.15 & 0.69 & 0.12 & 0.47\\ 
      & {u-Ticket} (ours) & \textbf{0.86} & \textbf{0.00} & \textbf{0.05} & \textbf{1.00}\\
\addlinespace[0.4em]
{CIFAR100} & LTH\cite{kim2022exploring} & 1.15 & 0.69 & 0.12 & 0.47\\ 
      & {u-Ticket} (ours) & \textbf{0.86} & \textbf{0.00} & \textbf{0.05} & \textbf{1.00}\\ 
\midrule
\multicolumn{6}{c}{{ResNet-19}\cite{resnet}} \\
\midrule

{CIFAR10}  & LTH\cite{kim2022exploring} & 1.66 & 1.73 & 0.21 &0.31 \\ 
         & {u-Ticket} (ours) & \textbf{1.10} & \textbf{0.00} & \textbf{0.07} & \textbf{1.00}\\
\addlinespace[0.4em]
{FMNIST}  & LTH\cite{kim2022exploring} & 1.26 & 1.89 & 0.20 & 0.27\\ 
        & {u-Ticket} (ours) & \textbf{0.73} & \textbf{0.00} & \textbf{0.05} & \textbf{1.00}\\
\addlinespace[0.4em]
{SVHN}  & LTH\cite{kim2022exploring} & 1.27 & 1.34 & 0.16 & 0.30\\ 
      & {u-Ticket} (ours) & \textbf{0.84} & \textbf{0.00} & \textbf{0.05} & \textbf{1.00}\\
\addlinespace[0.4em]
{CIFAR100} & LTH\cite{kim2022exploring} & 1.15 & 0.69 & 0.12 & 0.47\\ 
      & {u-Ticket} (ours) & \textbf{0.86} & \textbf{0.00} & \textbf{0.05} & \textbf{1.00}\\ 
\bottomrule
\end{tabular} 
\label{speed_tab}
\end{adjustbox}
\end{table}

\begin{figure}[t]
{\hspace{8mm}\includegraphics[width=0.8\linewidth]{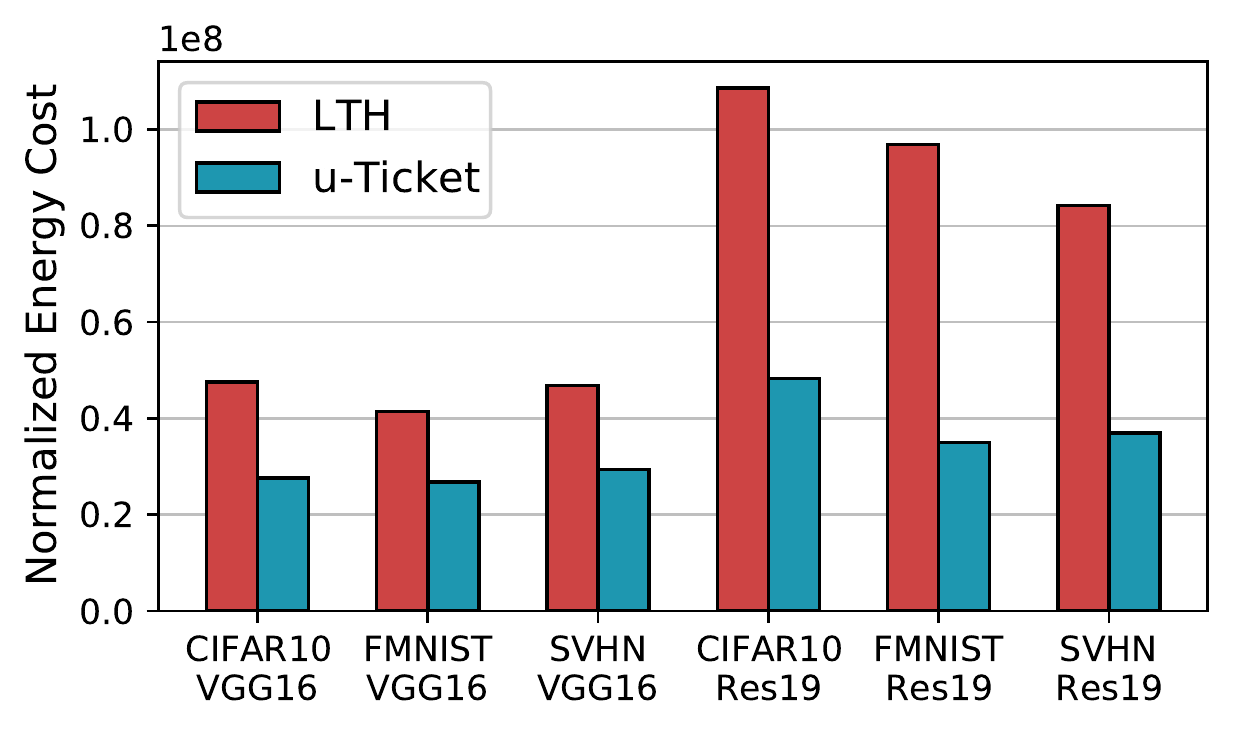}}
\vspace{-3mm}
\caption{Comparison of the normalized energy cost between two networks and across three datasets. The energy results are normalized to the energy required by a PE to process one input spike.}
\label{fig:energy_rsults}
\end{figure}

\subsubsection{Energy Performance}
In this section, we further show the energy efficiency improvements of u-Ticket over the standard LTH baseline. The energy differences are visualized in Fig. \ref{fig:energy_rsults} (a), from which we observe that the energy benefits of balancing the workloads are huge. For CIFAR10, FMNIST, and SVHN, we managed to reduce the energy cost by $41.8\%$, $35.4\%$, and $37.2\%$ on VGG-16, and $55.5\%$, $63.8\%$, and $56.1\%$ on ResNet-19.

The main source of energy cost reduction comes from eliminating idle cycles and reducing latency, which ultimately reduces the leakage energy of the hardware. ResNet-19, whose network is deeper, suffers more from the workload imbalance problem and thus has more idle cycles and longer latency compared to VGG-16. By eliminating almost all the idle cycles, u-Ticket brings more energy cost reduction to ResNet-19 compared to VGG-16.

\subsubsection{Searching and Recovering Speed}
\begin{table}[t!]
\centering
\caption{Comparison between the performance of u-Ticket with and without the early-ticket (ET) method.}
\vspace*{-1mm}
\begin{adjustbox}{max width =\linewidth}
\begin{tabular}{@{\extracolsep{4pt}}lcccc}
\toprule   
{Method} & {Acc.(\%)} & {Sparsity(\%)} & {Utilization} & Searching Time\\
 &  & {(filters)} & &(seconds/round)\\
\midrule
\multicolumn{5}{c}{{VGG-16}, CIFAR10} \\
\midrule
 LTH\cite{kim2022exploring} & \textbf{91.0} & 98.2 & 0.66 & 3031\\ 
 {u-Ticket}& 90.7 & \textbf{98.4} & \textbf{1.00} & 3032\\
 LTH ET~\cite{kim2022exploring} & 90.9 & 98.2 & 0.59 & 1553\\
 {u-Ticket ET}& 90.6 & \textbf{98.3} & \textbf{1.00}& 1558\\
\bottomrule
\end{tabular}
\label{et_tab}
\end{adjustbox}
\end{table}

Several techniques in other LTH-based work are used to reduce the searching time~\cite{kim2022exploring,you2019drawing}. We further apply the Early-Ticket~\cite{kim2022exploring} 
 (ET) to u-Ticket to reduce the ticket searching time for SNNs. As shown in the Table.~\ref{et_tab}, u-Ticket works well with ET. By applying ET, u-Ticket can still recover the utilization to 100\% at iso-accuracy and weight sparsity, with approximately 50\% searching time reduction. The result suggests that our proposed method is orthogonal to other existing techniques that reduce the searching time for standard LTH.

\subsection{Ablation Studies}

\begin{figure}[t]
\begin{center}
\def\arraystretch{0.5}
\begin{tabular}{@{}c@{}c}
\includegraphics[width=0.5\linewidth]{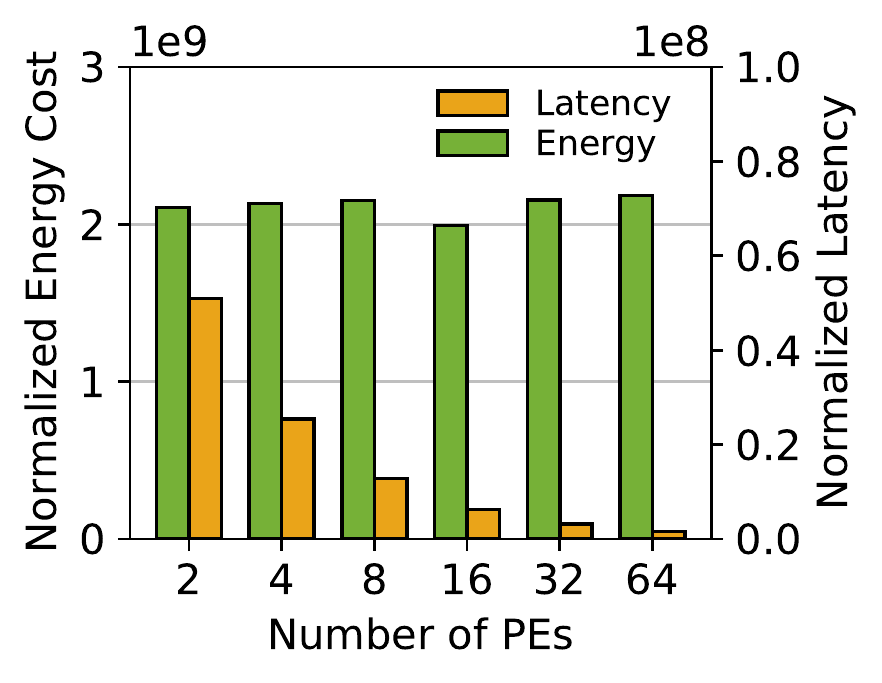} &
\includegraphics[width=0.48\linewidth]{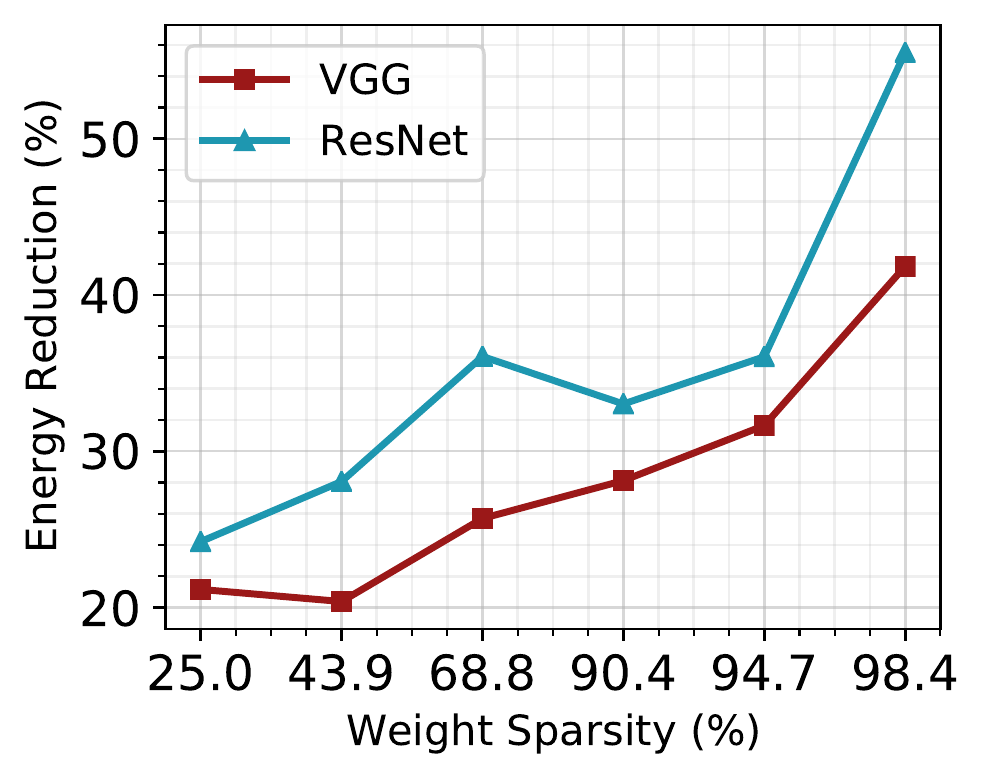} 
\\
{ (a)} & { (b)} 
\end{tabular}
\caption{(a) Comparison of the normalized energy cost between two networks across three datasets. The energy results are normalized to the energy required by a PE to process one input spike.
(b) Percentage of normalized energy reduction compared to the LTH baseline for different weight sparsity.
}\vspace{-3mm}
\label{fig:energy_rsults_ablation}
\end{center}
\end{figure}

\noindent\textbf{Analysis of Sparsity:} We study the effects of the u-Ticket method under different weight sparsity. We measure the energy difference between u-Ticket and the LTH baseline at different pruning rounds for both ResNet-19 and VGG-16 on the CIFAR10 dataset. The result is visualized in Fig. \ref{fig:energy_rsults_ablation} (b). As observed, with increased weight sparsity, the benefit of using u-Ticket gets greater. This is due to the degradation of the utilization in LTH as aforementioned in Fig. \ref{lth_problem}.

\noindent\textbf{Analysis of \#PEs:} We further study the effects of changing the number of PEs. We run the u-Ticket for VGG-16 on CIFAR10 with 2, 4, 8, 16, 32, and 64 PEs, illustrating the results in Fig. \ref{fig:energy_rsults_ablation} (a). The latency decreases linearly while the energy cost changes slightly with an increasing number of PEs. Considering that the area of PE arrays will also linearly increase with the number of PEs, we conduct most of our experiments with 16 PEs, which is a suitable tradeoff point.

\begin{figure}[t]
\centerline{\includegraphics[width=0.9\linewidth]{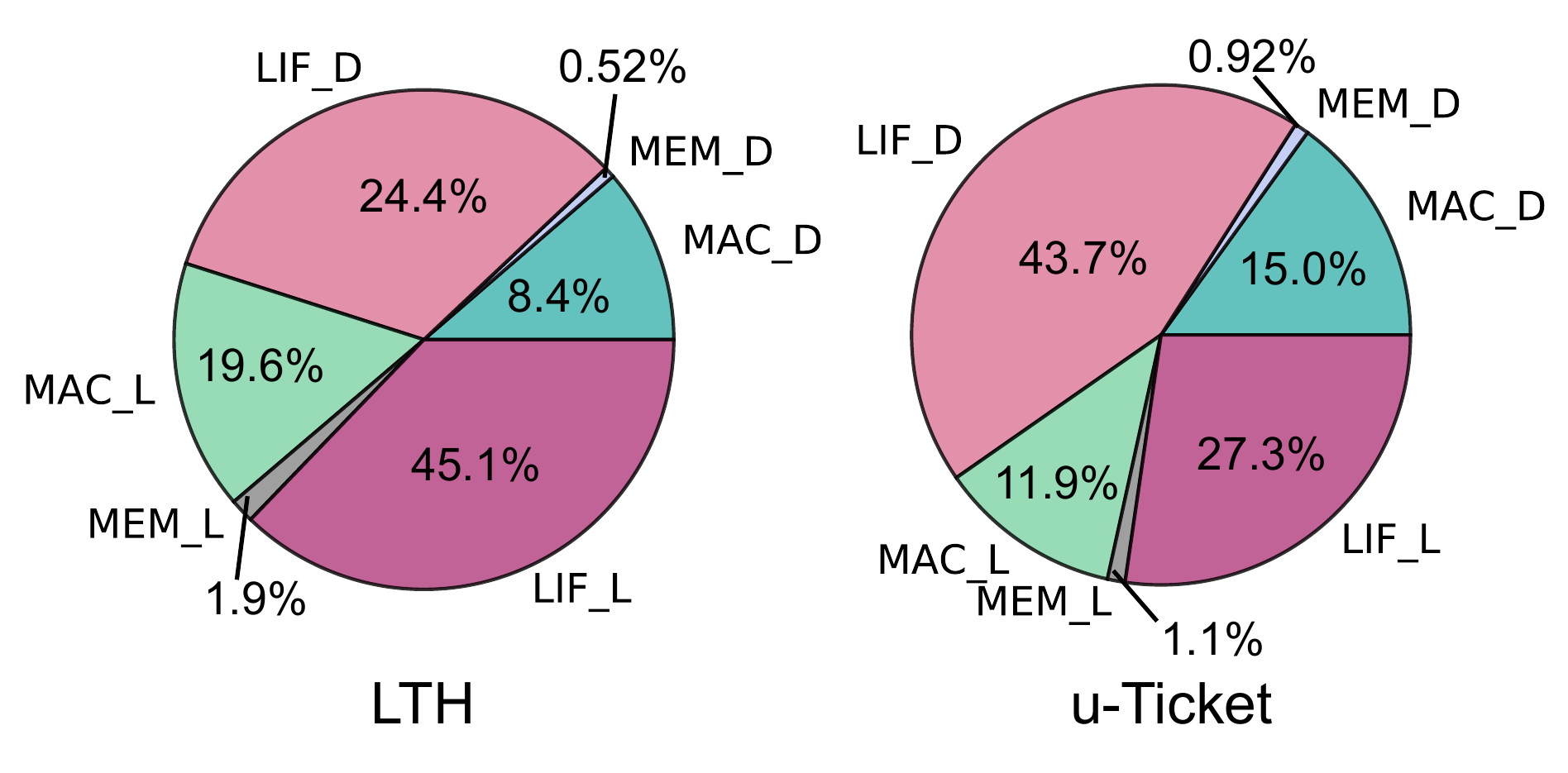}}
\vspace{-3mm}
\caption{Comparison of the energy breakdown between u-Ticket and the LTH baseline. MAC\_L, MEM\_L, and LIF\_L denote the leakage energy for MAC, LIF, and memory operation, while MAC\_D, MEM\_D, and LIF\_D denote their dynamic energy.}
% \vspace*{-2em}
\label{breakdown}
\end{figure}

\noindent\textbf{Analysis of Energy Breakdown:} In Fig. \ref{breakdown}, we show the energy breakdown comparison between u-Ticket and the LTH baseline on ResNet-19 for the CIFAR10 dataset. The energy components are the dynamic and leakage energy of MAC, LIF, and MEM operations (reading of SRAM-based scratchpad). We observe that the leakage energy for both MAC and LIF operation is significantly reduced in u-Ticket due to the elimination of idle cycles. Expectedly, the portion of the dynamic energy of MAC and LIF operation increases.

\begin{figure}[t]
\begin{center}
\def\arraystretch{0.5}
\begin{tabular}{@{}c@{}c}
\includegraphics[width=0.52\linewidth]{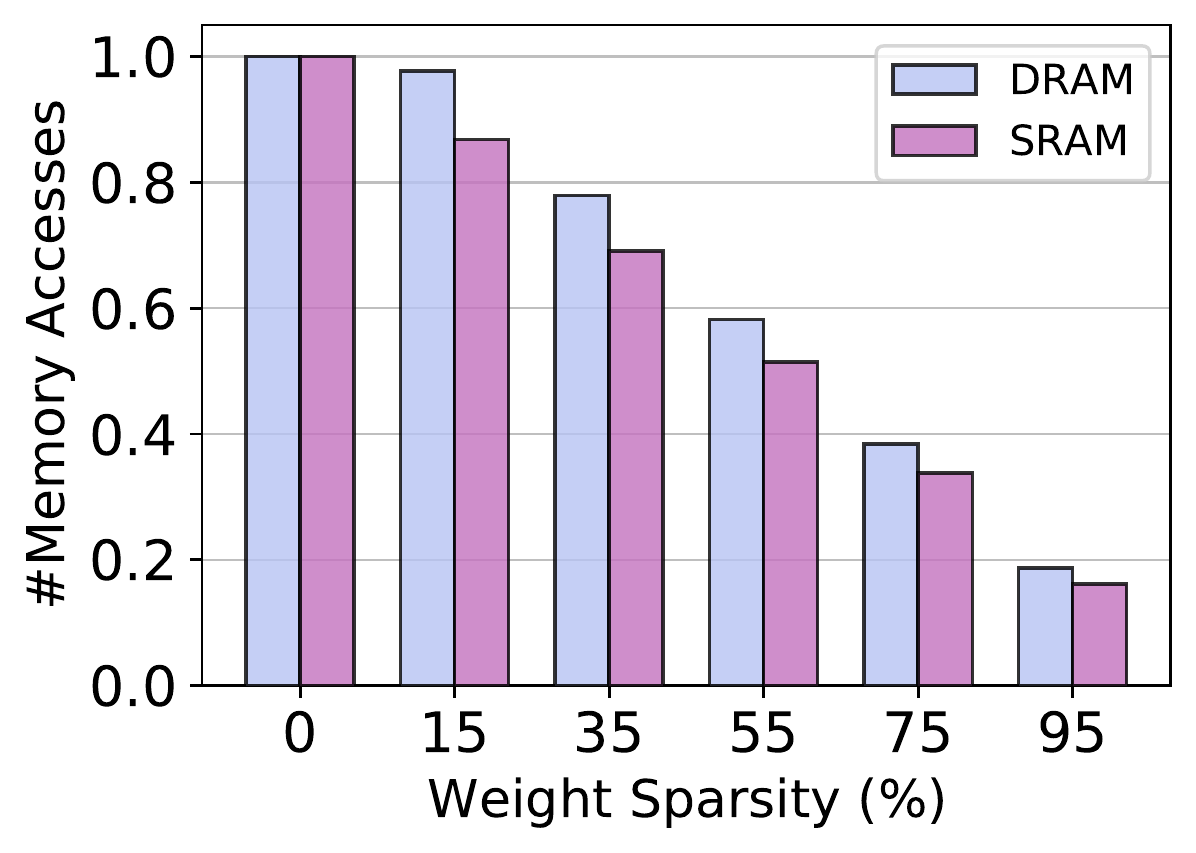} &
\includegraphics[width=0.48\linewidth]{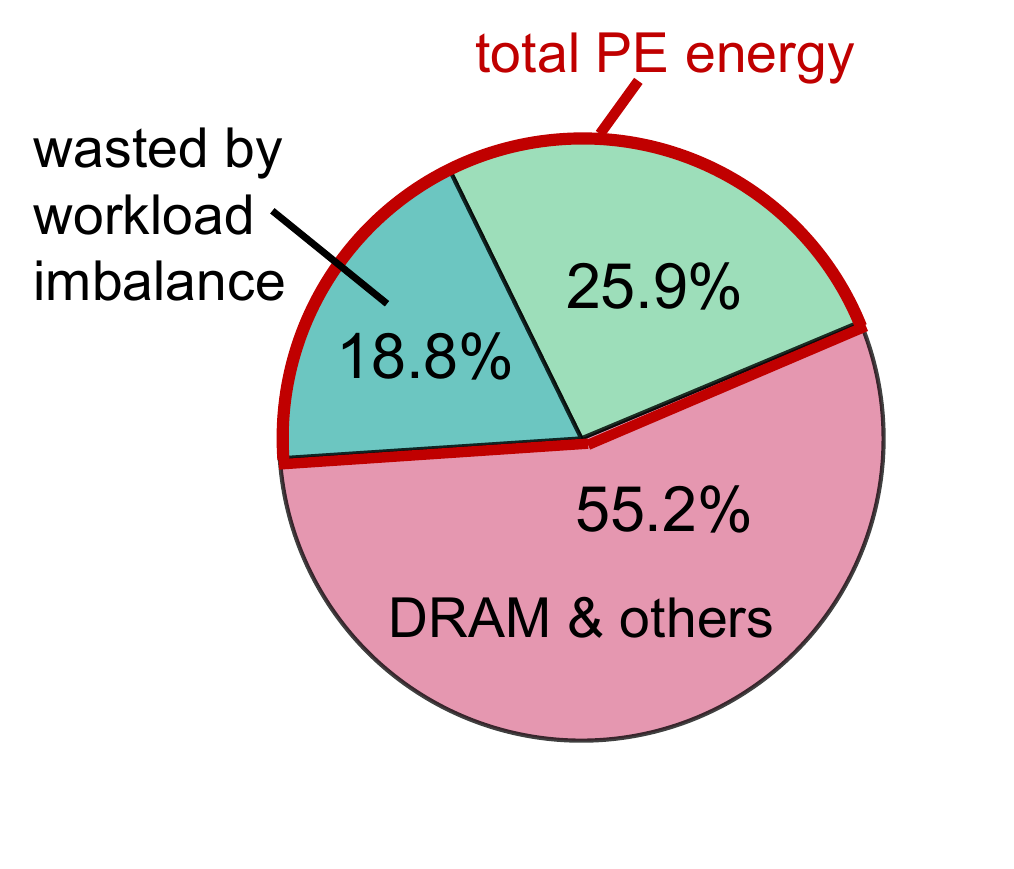} 
\\
{ (a)} & { (b)} 
\end{tabular}
\vspace{-3mm}
\caption{(a) Normalized DRAM and SRAM accesses comparison across different weight sparsity. (b) The component breakup of the total energy for LTH baseline with 95\% sparsity. Both results are shown for VGG-16 with CIFAR10.
}

\label{fig:mem_comp}
\end{center}
\end{figure}

\begin{wrapfigure}{r}{0.4\columnwidth}
 \centering
 \vspace{-3mm}
\includegraphics[width=0.4\columnwidth]{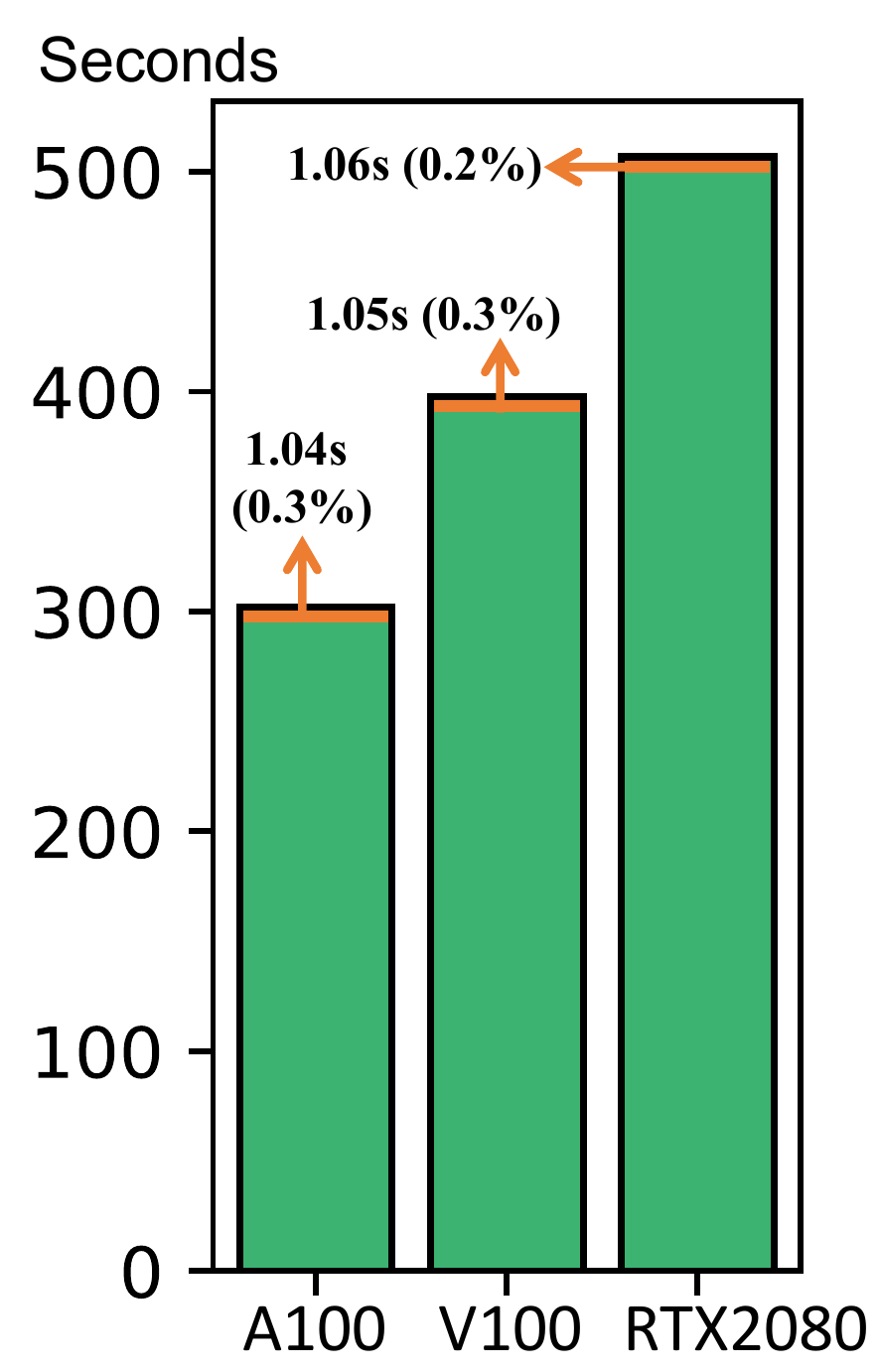}
\vspace{-7mm}
\caption{GPU latency of u-Ticket.}
\label{fig:gpu}
 \vspace{-2mm}
\end{wrapfigure}

\noindent\textbf{System Level Study:} 
Furthermore, we study the behavior of the overall system of sparse SNNs. In Fig. \ref{fig:mem_comp} (a), we show how the total DRAM and SRAM access (normalized with respect to the dense SNN) decrease with increasing weight sparsity. The results again encourage the necessity of pruning the networks into the extremely high sparsity domain. Furthermore, we find that in the extremely high weight sparsity regime, the PE level energy takes a significant portion of the total energy ($\sim$ 45\% on VGG-16 with CIFAR10). As a result, after applying u-Ticket to balance the PE workloads, we reduced approximately $19\%$ of the total energy at the system level, as shown in Fig.~\ref{fig:mem_comp}~(b).

\noindent\textbf{Recovery Overheads on GPUs: }
We also quantify the latency overheads of utilization check and recovery on multiple GPU devices. The result shows that our u-Ticket brings almost no latency overheads to the standard LTH across three CUDA GPUs: RTX-2080Ti, V100, and A100. The result is shown in Fig.~\ref{fig:gpu}. The latency is the total searching time of one complete u-Ticket search round. The green portion is the time for training on VGG-16 for the CIFAR10 dataset for 15 epochs. The orange part is the utilization recovery and check part (Line 6-16 in Algorithm~\ref{alg:u_rec}). Compared to the training time, u-Ticket's utilization check and recovery time is negligible ($\sim$ 0.3\% of one complete ticket searching time).

\noindent\textbf{Evaluation on Temporal-Series Datasets: } To further validate our method on datasets that heavily rely on temporal-series information, we conduct experiments using Dynamic Vision Sensor (DVS) datasets obtained from event-based cameras. Descriptions of these datasets are provided below:

\textbf{CIFAR10-DVS:} CIFAR10-DVS dataset~\cite{dvscifar10} contains 10K DVS images recorded from the CIFAR10 dataset~\cite{krizhevsky2009learning}. We resize the image resolution to 48 x 48 and divide the event series into 10 frames per image sample.

\textbf{N-Caltech 101:} N-Caltech 101 dataset~\cite{ncaltech} contains 8831 DVS images recorded from the Caltech 101 dataset. Similar to the CIFAR10-DVS, we resize the image resolution to 48 x 48 and divide the event data into 10 frames per sample. We run both datasets on the ResNet-19 network with 10 rounds of u-Ticket search. At each round, 25\% weights are pruned. 

\begin{figure}[t]
\centerline{\includegraphics[width=0.8\linewidth]{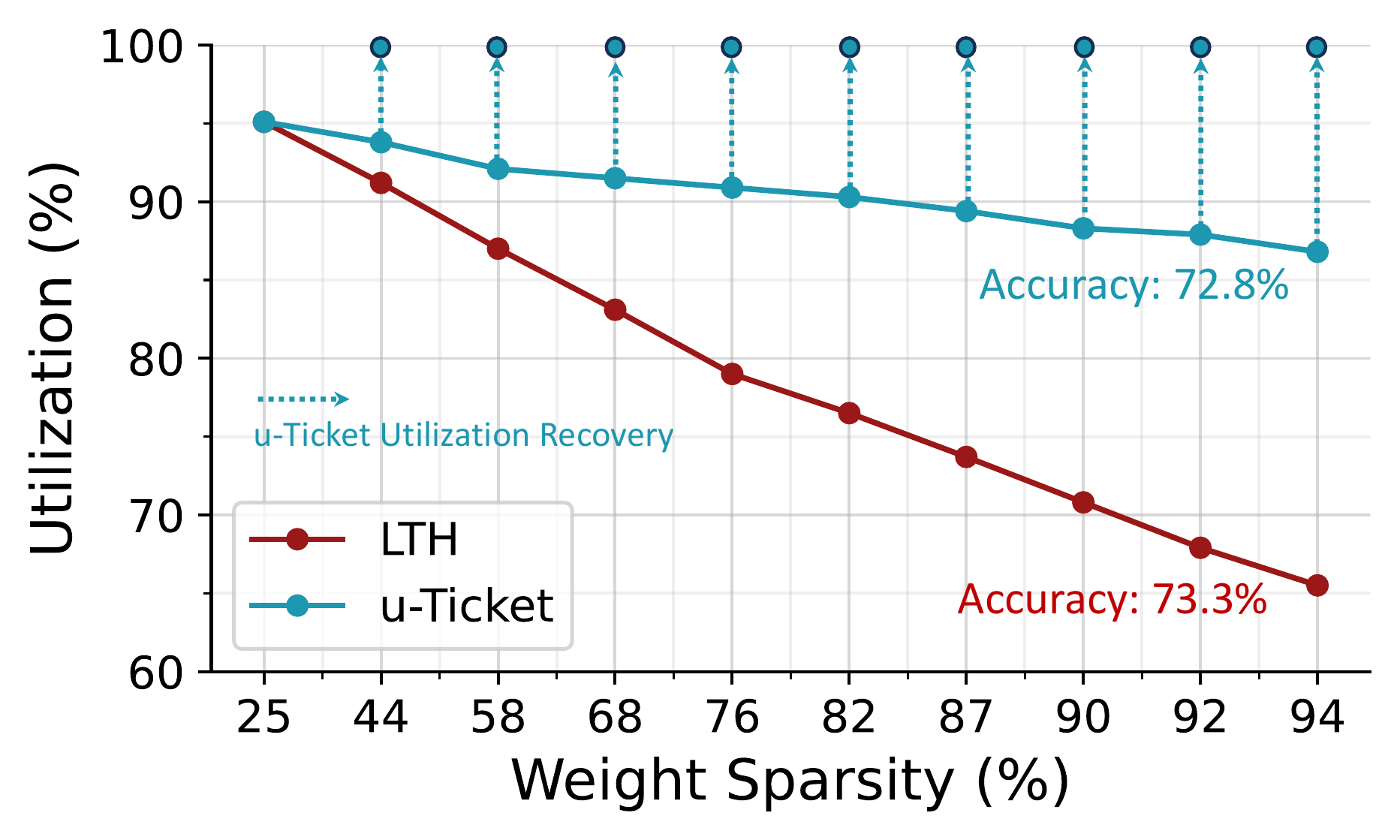}}
\caption{Weight sparsity vs. utilization for CIFAR10-DVS on ResNet-19.}
% \vspace*{-2em}
\label{dvs}
\end{figure}

\begin{table}[t!]

\centering

\caption{Ablation study: comparison of accuracy, sparsity of filters, and utilization between u-Ticket and LTH method on DVS datasets.}
\vspace*{-1mm}
\begin{adjustbox}{max width =\linewidth}
\begin{tabular}{@{\extracolsep{4pt}}llccc}
\toprule   
 {Dataset} & {Method} & {Acc.(\%)} & {Sparsity(\%)} & {Utilization}\\
 &  &  & {(filters)} & \\
\midrule
\multicolumn{5}{c}{{ResNet-19}\cite{resnet}} \\
\midrule
{CIFAR10-DVS}  & LTH\cite{kim2022exploring} & \textbf{73.3} & 94.4 & 0.66\\ 
         & {u-Ticket} (ours) & 72.8 & \textbf{95.3} & \textbf{1.00}\\
\addlinespace[0.4em]
{N-Caltech101}  & LTH\cite{kim2022exploring} & 63.2 & \textbf{98.2} & 0.54\\ 
        & {u-Ticket} (ours) & \textbf{64.9} & 97.8 & \textbf{1.00}\\
\bottomrule
\end{tabular}
\label{dvs_tab}
\end{adjustbox}
\end{table}
We validate our u-Ticket method in Table.~\ref{dvs_tab}. For both the DVS datasets, our method achieves iso-accuracy compared to the original LTH method with slightly higher weight sparsity and 100\% utilization. We further illustrate the trend of utilization and weight sparsity on CIFAR10-DVS in Fig.~\ref{dvs}. Our method yields a ticket with better utilization in every round. Thus, we can easily recover the utilization to 100\% for that ticket without accuracy degradation.

\section{Discussion}
\subsection{Comparison with Structured Pruning}
In this work, we target to solve the workload imbalance problem associated with unstructured pruning. In contrast to unstructured pruning, structured pruning has also been a popular network compression method~\cite{cascade,lss,nmprun}. In structured pruning, the networks are pruned in a pattern that aims to leverage the hardware's power to process the pruned networks more efficiently. The nature of the structured pruning does not make it suffer from the workload imbalance problem we discussed in this work. Although the structured pruning methods take advantage of efficient hardware processing, they do suffer from a relatively lower weight sparsity. For example, on the VGG-16 network, the structured pruning, on average, achieves around 85\% weight sparsity, while our LTH-based unstructured pruning gets over 95\% weight sparsity.

\subsection{Compatibility on Async Neuromorphic Chip}
While in the paper, we have limited our discussion of the uTicket's hardware benefits to the synchronized digital accelerators. It is worth noting that our method also has the potential to improve the utilization of the async neuromorphic chips~\cite{akopyan2015truenorth,davies2018loihi,speck} when deploying the LTH-based SNN models. Assume that a sparse SNN is deployed on the neuromorphic chip. Depending on the number of synaptic connections, different post-synaptic neurons will receive different numbers of event-driven packages between timestep $t$ to $t+1$. This will result in an imbalanced processing time across different cores. Furthermore, as a popular design choice in neuromorphic chips, every time the chip advances its timestep, each core will have a barrier synchronization~\cite{barrier}. Consequently, the heavily imbalanced sparse networks will lead to a longer waiting time for the idling cores during the barrier synchronization process. So, in conclusion, as long as the neuromorphic chips take some synchronization steps between the asynchronized computations, there will be a workload imbalance problem. Our method can potentially provide a workload balance solution without any hardware modifications on the chip. Moreover, this discussion should also apply to other chip designs that use addressing algorithms.

\subsection{Future Direction}
We suggest several interesting potential future directions based on this work. Firstly, although the u-Ticket method works well on recovering the utilization at iso-accuracy and iso-weight-sparsity, the observations are based on empirical experiment results. A detailed analysis of the mathematical reason behind this would benefit the community. Moreover, although the experiment results show that our workload balancing method would not hurt the sparse firing activity of SNNs, theoretically studying the relationship between the workload utilization and the spike firing activity of SNNs is important. We envision that the probability model from~\cite{yao2023probabilistic}, which builds the relationship between pruning and SNN firing, would be a good starting point.
Further, this work is motivated by the hardware-resource limitation of SNNs; thus, we focus our experiments and analysis on SNNs. However, whether our method applies to ANNs can be another useful insight for the community.

\section{Conclusion}
In this work, we propose u-Ticket, a utilization-aware LTH-based pruning method that solves the workload imbalance problem in SNNs. 
Unlike prior works, u-Ticket recovers the utilization during pruning, thus avoiding additional hardware to balance the workloads during deployment. Additionally, at iso-accuracy, u-Ticket improves PE utilization by up to 100\% compared to the standard LTH-based pruning method while maintaining filter sparsity of 98\%. Moreover, u-Ticket reduces the running latency by up to 77\% and energy cost by up to 64\% compared to the standard LTH baseline.

\section{Acknowledgements}

This work was supported in part by CoCoSys, a JUMP2.0 center sponsored by DARPA and SRC, Google Research Scholar Award, the National Science Foundation CAREER Award, TII (Abu Dhabi), the DARPA AI Exploration (AIE) program, and the DoE MMICC center SEA-CROGS (Award \#DE-SC0023198).

\bibliographystyle{IEEEtran}
\bibliography{main}

\begin{IEEEbiography}[{\includegraphics[width=1in,height=1.25in, clip,keepaspectratio]{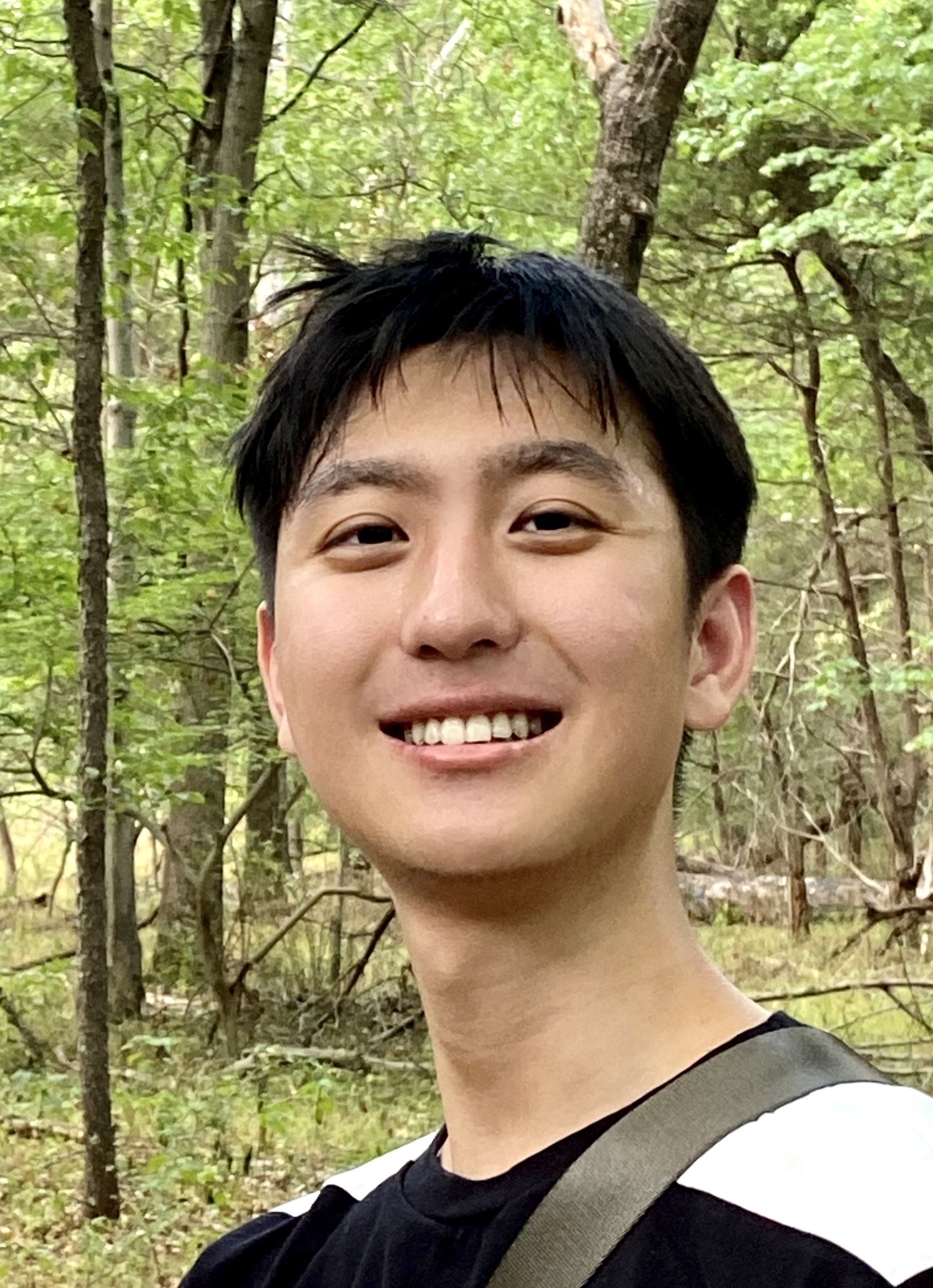}}]{Ruokai Yin} is a Ph.D. student in the Department of Electrical Engineering at Yale University. His research focuses on various computing schemes, with a particular emphasis on neuromorphic computing, as enablers for energy-efficient machine learning. He is especially interested in designing low-power computer architectures and systems tailored to these computing schemes. Additionally, he works on hardware-aware compression algorithms for neural networks, specifically aimed at enhancing energy efficiency during deployment. Prior to joining Yale, he earned his B.S. in Electrical Engineering from the University of Wisconsin-Madison.
\end{IEEEbiography}

\begin{IEEEbiography}[{\includegraphics[width=1in,height=1.25in, clip,keepaspectratio]{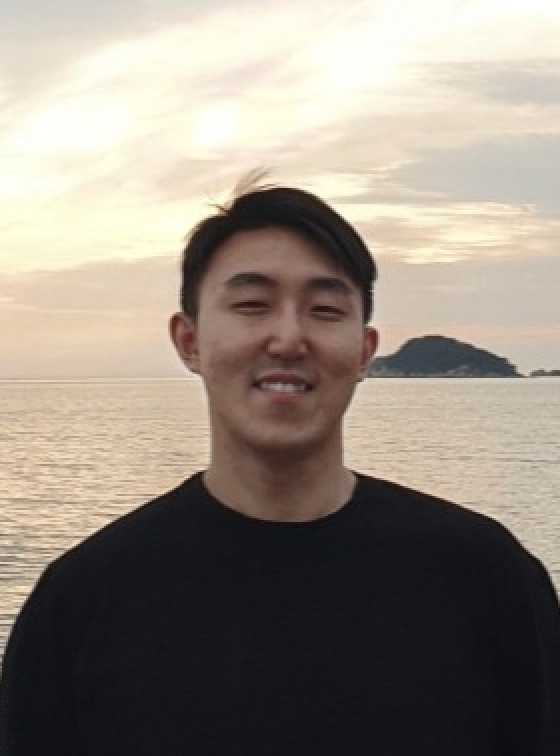}}]{Youngeun Kim} is currently working toward a Ph.D. degree in Electrical Engineering at Yale University, New Haven, CT, USA.
Prior to joining Yale,  he worked as a full-time student intern at T-Brain, AI Center, SK telecom, South Korea. He received his B.S. degree in Electronic Engineering from Sogang University, South Korea, in 2018 and M.S. degree in Electrical Engineering from Korea Advanced Institute of Science and Technology (KAIST), in 2020. His research interests include neuromorphic computing, computer vision, and deep learning.
\end{IEEEbiography}

\begin{IEEEbiography}[{\includegraphics[width=1in,height=1.25in, clip,keepaspectratio]{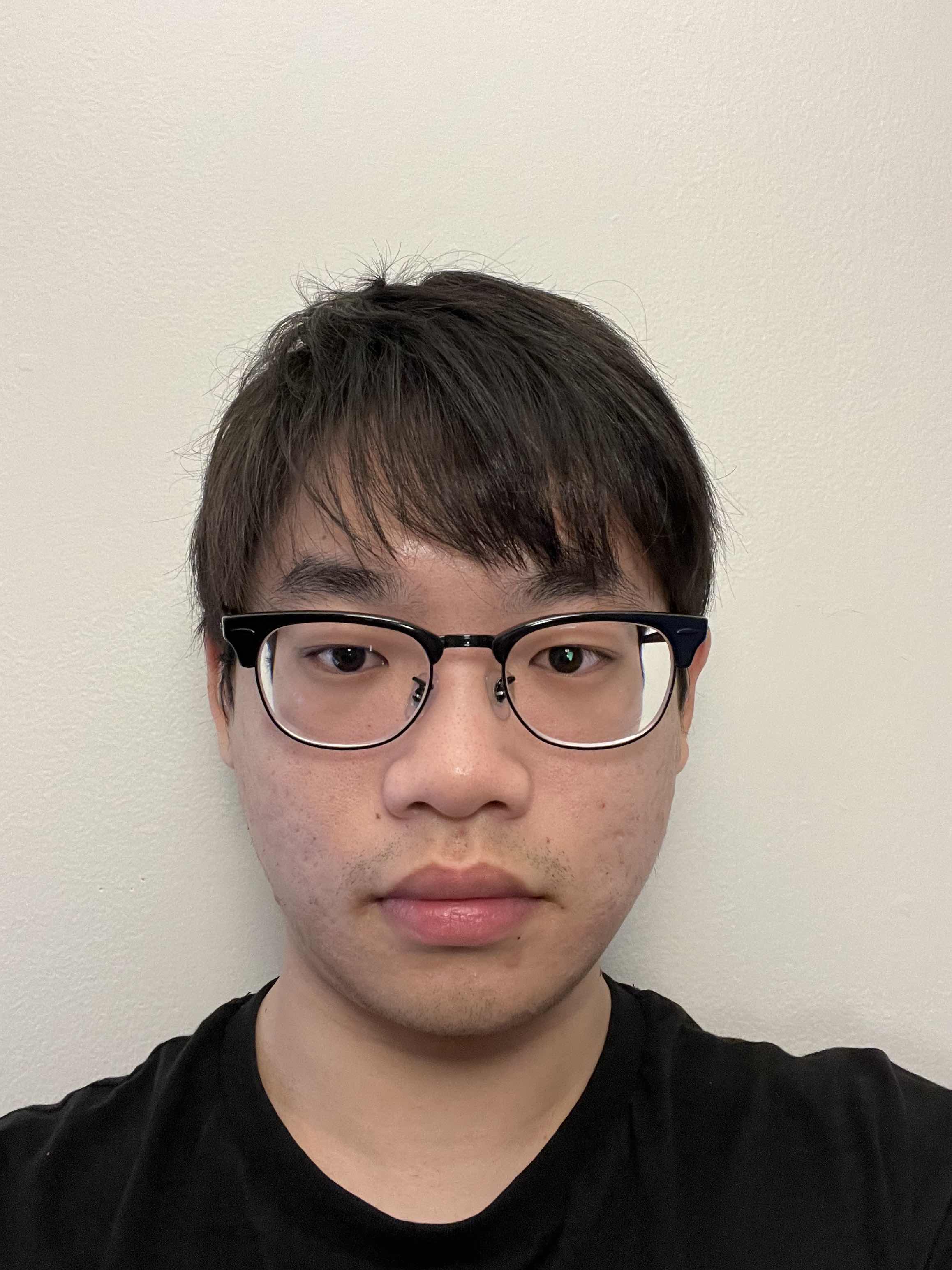}}]{Yuhang Li} received the B.E. in Department of Computer Science and Technology, University of Electronic Science and Technology of China (UESTC) in 2020. He was a research assistant at the National University of Singapore and UESTC in 2019 and 2021, respectively. Now he is pursuing his Ph.D. degree at Yale University, supervised by Prof. Priyadarshini Panda. His research interests include Efficient Deep Learning, Brain-inspired Computing, and Model Compression. 
\end{IEEEbiography}

\begin{IEEEbiography}
[{\includegraphics[width=1in,height=1.25in, clip,keepaspectratio]{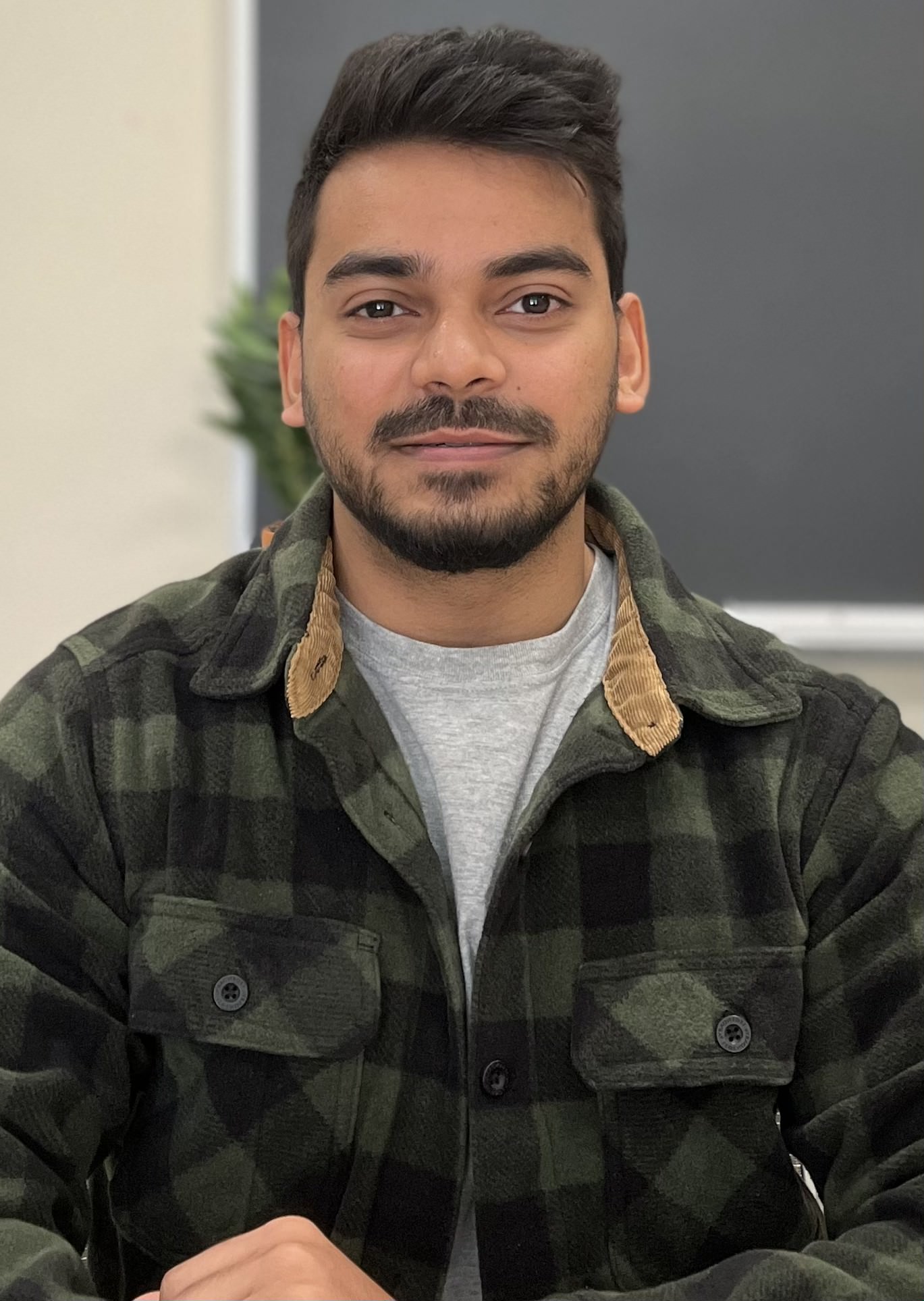}}]{Abhishek Moitra} is pursuing his Ph.D. in the Intelligent Computing Lab at Yale. His research works have been published in reputed journals such as IEEE TCAS-1, IEEE TCAD and conferences such as DAC. His research interests involve hardware-algorithm co-design and co-exploration for designing robust and energy-efficient hardware architectures for deep learning tasks.
\end{IEEEbiography}

\begin{IEEEbiography}
[{\includegraphics[width=1in,height=1.25in, clip,keepaspectratio]{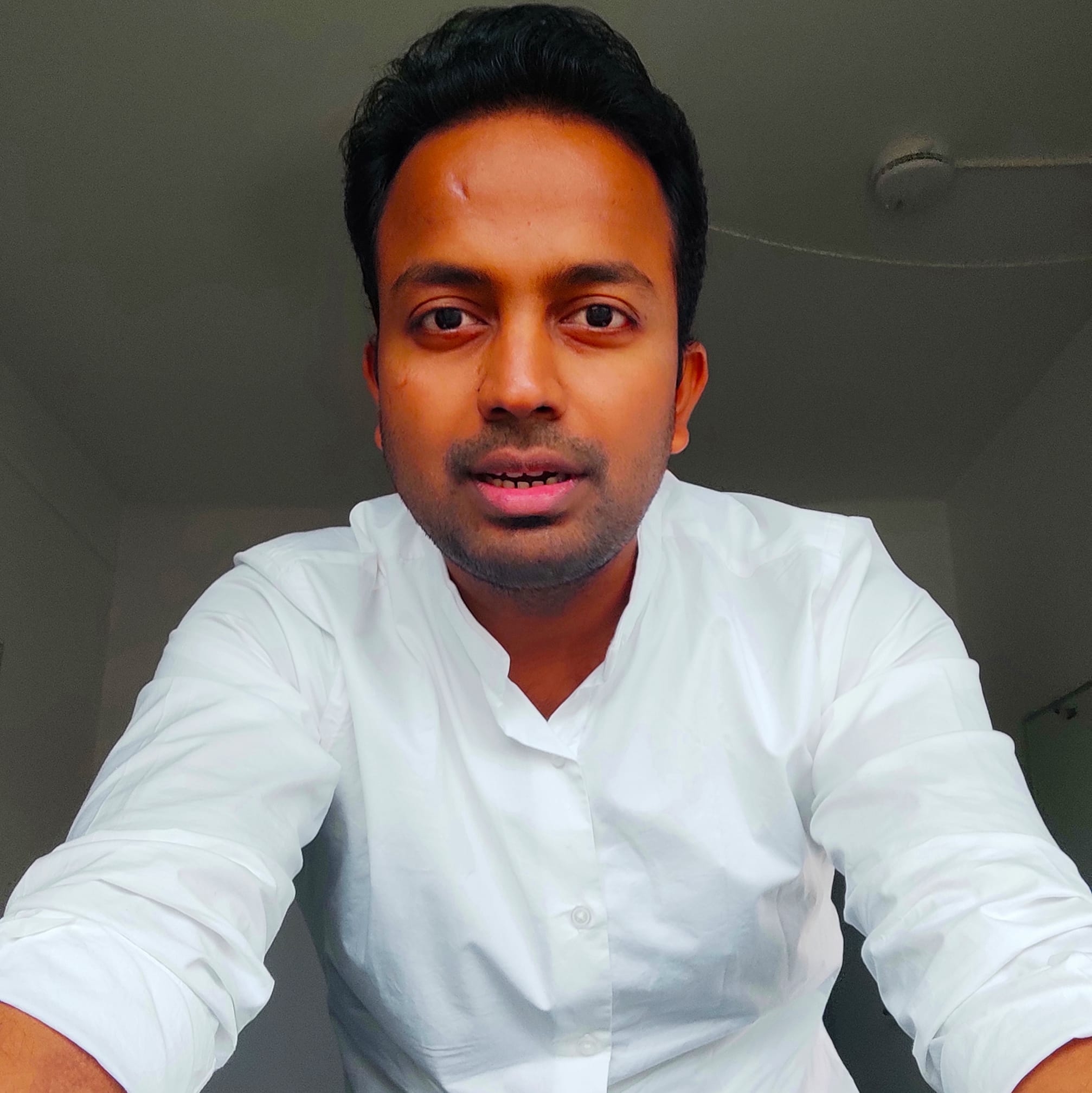}}]{Nitin Satpute} is a Senior Machine Learning Engineer at Technology Innovation Institute Abu Dhabi, UAE. He has done his PhD from University of Cordoba, Spain and Master of Engineering (ME) from Birla Institute of Technology and Science (BITS) Pilani, India. He is currentyly involved in the development and implementation of Distribuited AI for Cryptography. His research interests include AI, HPC, Biomedical Imaging, Computer Vision and Cryptography.
\end{IEEEbiography}

\begin{IEEEbiography}
[{\includegraphics[width=1in,height=1.25in, clip,keepaspectratio]{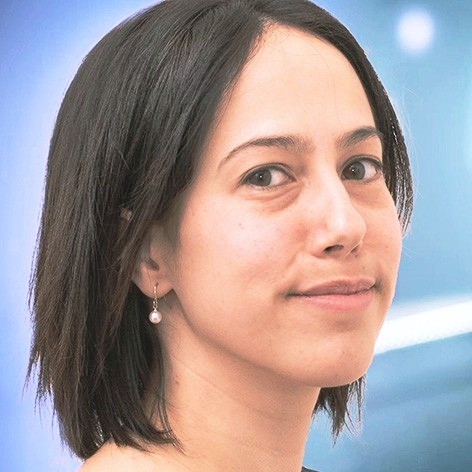}}]{Anna Hambitzer} is a Senior Data Scientist at the Cryptography Research Center, TII, Abu Dhabi. She holds a Ph.D. in Information Technology from ETH Zurich, and has over five years of experience in AI and data analytics, including two years in the financial industry. 
\end{IEEEbiography}

\begin{IEEEbiography}[{\includegraphics[width=1in,height=1.25in, clip,keepaspectratio]{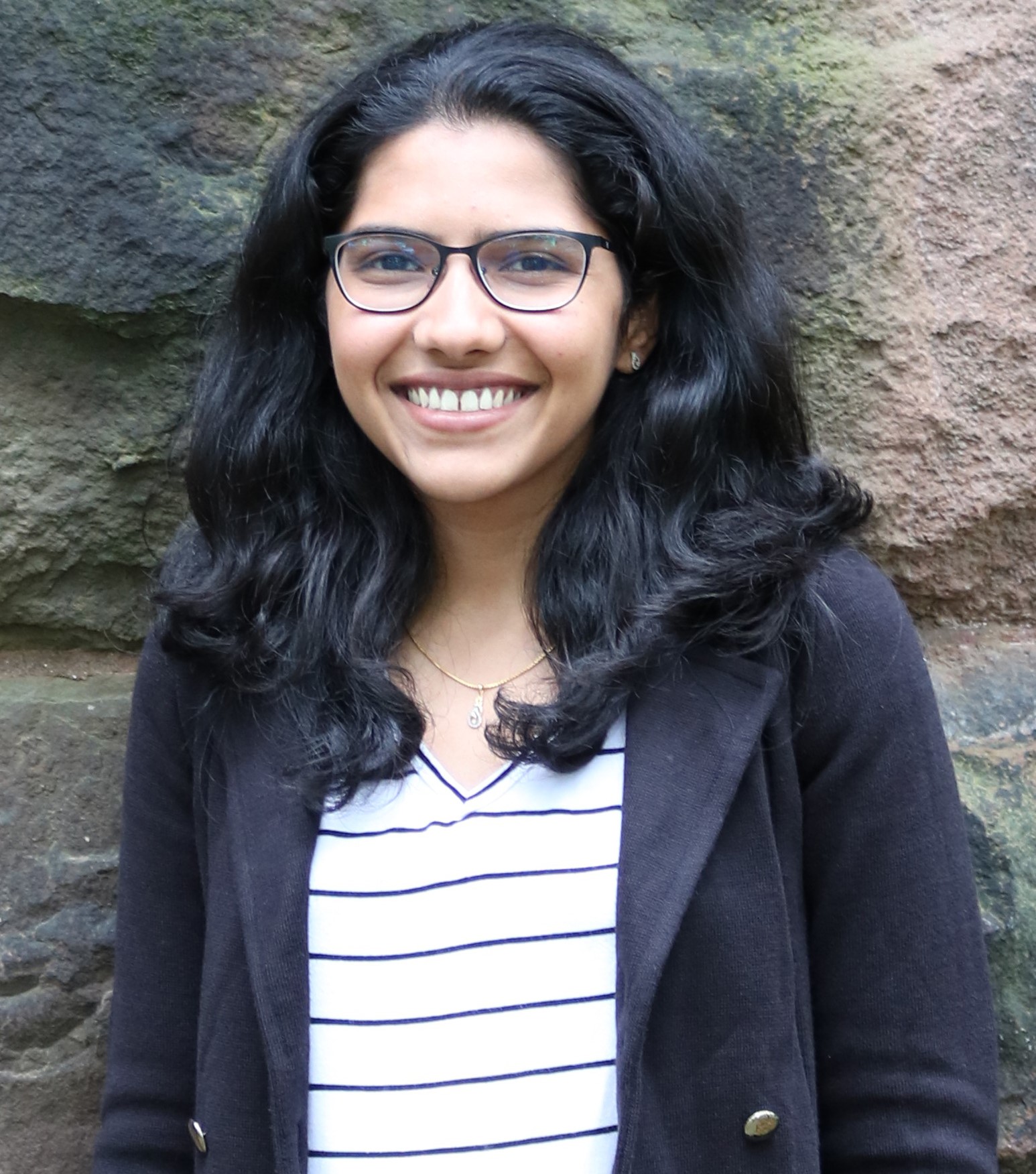}}]{Priyadarshini Panda} is an assistant professor in the electrical engineering department at Yale University, USA. She received her B.E. degree in Electrical \& Electronics and Master's degree in Physics from BITS, Pilani, India in 2013 and her Ph.D. in Electrical \& Computer Engineering from Purdue University, USA in 2019. She was the recipient of outstanding student award in Physics in 2013. In 2017, she interned at Intel Labs, Oregon, USA where she developed large-scale spiking neural network algorithms for benchmarking the Loihi chip. She is the recipient of the 2019 Amazon Research Award, 2022 Google Scholar Research Award, and 2022 DARPA Riser Award. She has published more than 60 publications in well-recognized venues including, Nature, Nature Communications, and IEEE among others. Her research interests include- neuromorphic computing, energy-efficient deep learning, adversarial robustness, and hardware-centric design of robust neural systems.
\end{IEEEbiography}
\vfill

\end{document}